\documentclass{article}
\usepackage[preprint]{neurips_2026}
\usepackage{amsmath,amssymb,amsthm}
\usepackage{graphicx,booktabs,enumitem,xspace,multirow,array,tabularx,longtable}
\usepackage[T1]{fontenc}
\usepackage[utf8]{inputenc}
\usepackage{microtype}
\usepackage{xcolor}
\usepackage[most]{tcolorbox}
\usepackage{hyperref}
\hypersetup{hidelinks,
  pdftitle={What Can Activation Patching Actually Establish? Reachable-State Geometry and the Limits of Mechanistic Evidence},
  pdfauthor={Ankush Kadu, Ananth Kalyanasundaram, Aswanth Krishnan},
  pdfkeywords={activation patching, mechanistic interpretability, identifiability, experimental design, evaluation}}
\usepackage{xurl}
\usepackage{placeins}

\newcommand{\AdqCentreJ}{0.00232\xspace}  % adq.J.centre: median over capture
\newcommand{\AdqCentreS}{0.169\xspace}  % adq.S.centre: median over capture
\newcommand{\AdqCoverIntervalJ}{0/31\xspace}  % adq.d3.J.interval_all: count over capture
\newcommand{\AdqCoverIntervalS}{3/31\xspace}  % adq.d3.S.interval_all: count over capture
\newcommand{\AdqCoverUpperJ}{1/31\xspace}  % adq.d3.J.upper_all: count over capture
\newcommand{\AdqCoverUpperJFour}{2/31\xspace}  % adq.d4.J.upper_all: count over capture
\newcommand{\AdqCoverUpperJTwo}{0/31\xspace}  % adq.d2.J.upper_all: count over capture
\newcommand{\AdqCoverUpperS}{12/31\xspace}  % adq.d3.S.upper_all: count over capture
\newcommand{\AdqCoverUpperSFour}{15/31\xspace}  % adq.d4.S.upper_all: count over capture
\newcommand{\AdqCoverUpperSTwo}{11/31\xspace}  % adq.d2.S.upper_all: count over capture
\newcommand{\AdqEffectAlphaOne}{0.322\xspace}  % adq.effect_alpha1: median over capture
\newcommand{\AdqEnvJ}{0.0129\xspace}  % adq.J.env.median: median over captures over capture
\newcommand{\AdqEnvS}{0.511\xspace}  % adq.S.env.median: median over captures over capture
\newcommand{\AdqEnvSCluster}{36.9\xspace}  % adq.cov.S_cr1_donor: median of per-capture percent over capture
\newcommand{\AdqEnvSHCzero}{27.6\xspace}  % adq.cov.S_hc0: median of per-capture percent over capture
\newcommand{\AdqEnvSOverRmsT}{72.1\xspace}  % adq.envS_over_rmsT: ratio of medians over capture
\newcommand{\AdqIndRejJTwo}{0/18\xspace}  % adq.ind.rejJ.2way: count over configuration
\newcommand{\AdqIndRejSHi}{10\xspace}  % adq.ind.rejS.hi: max over robust estimators over configuration
\newcommand{\AdqIndRejSHomo}{1/18\xspace}  % adq.ind.rejS.homo: count over configuration
\newcommand{\AdqIndRejSLo}{7\xspace}  % adq.ind.rejS.lo: min over robust estimators over configuration
\newcommand{\AdqMaxTPctEffect}{17.4\xspace}  % adq.maxT_pct_effect: median of per-capture ratios over capture
\newcommand{\AdqRadiusJ}{0.0102\xspace}  % adq.J.radius: median over capture
\newcommand{\AdqRadiusS}{0.337\xspace}  % adq.S.radius: median over capture
\newcommand{\AdqRmsT}{0.00709\xspace}  % adq.rmsT.median: median over captures over capture
\newcommand{\AdqRmsTHi}{0.0133\xspace}  % adq.rmsT.q75: q75 over captures over capture
\newcommand{\AdqRmsTLo}{0.00472\xspace}  % adq.rmsT.q25: q25 over captures over capture
\newcommand{\AdqRmsTPct}{0.427\xspace}  % adq.rmsTpct.median: median over captures over capture
\newcommand{\AdqRmsTPctEffect}{2.62\xspace}  % adq.rmsT_pct_effect: median of per-capture ratios over capture
\newcommand{\AdqSdBetweenPrompt}{0.438\xspace}  % adq.disp.between: sd over prompt
\newcommand{\AdqSdWithinPrompt}{0.682\xspace}  % adq.disp.within: pooled sd over capture
\newcommand{\AdqSeedOneEnvS}{15.6\xspace}  % adq.seed1.envS: median within prompt over capture
\newcommand{\AdqSeedOneOverOne}{1/10\xspace}  % adq.seed1.overone: count over capture
\newcommand{\AdqSeedTwoEnvS}{21.7\xspace}  % adq.seed2.envS: median within prompt over capture
\newcommand{\AdqSeedTwoOverOne}{1/9\xspace}  % adq.seed2.overone: count over capture
\newcommand{\AdqSeedZeroEnvS}{146\xspace}  % adq.seed0.envS: median within prompt over capture
\newcommand{\AdqSeedZeroOverOne}{7/12\xspace}  % adq.seed0.overone: count over capture
\newcommand{\AdqSwingOverEffect}{6.47\xspace}  % adq.swing_over_effect: median of per-capture ratios over capture
\newcommand{\AngleAfter}{1.21\xspace}  % reparam.angle_after: median of within-cell medians of per-capture median angles over cell
\newcommand{\AngleBefore}{68.7\xspace}  % reparam.angle_before: median of within-cell medians of per-capture median angles over cell
\newcommand{\AuditDonorDupCaptures}{72/185\xspace}  % audit.donor_dup: count over capture
\newcommand{\AuditXseedSamePrompt}{10/120\xspace}  % audit.xseed.same_prompt: count over capture
\newcommand{\BGPTFourierSblind}{1295\xspace}  % basis.GPT-2.fourier.S: median over capture
\newcommand{\BGPTPolySblind}{1.36\times10^{7}\xspace}  % basis.GPT-2.poly.S: median over capture
\newcommand{\BGPTRbfSblind}{7.01\times10^{8}\xspace}  % basis.GPT-2.rbf.S: median over capture
\newcommand{\BQwenFourierSblind}{804\xspace}  % basis.Qwen3.5.fourier.S: median over capture
\newcommand{\BQwenPolySblind}{7.51\times10^{4}\xspace}  % basis.Qwen3.5.poly.S: median over capture
\newcommand{\BQwenRbfSblind}{7.91\times10^{6}\xspace}  % basis.Qwen3.5.rbf.S: median over capture
\newcommand{\CoordDiffSblind}{8.04\times10^{7}\xspace}  % coord.diff.S: median over capture
\newcommand{\CoordPConeSblind}{7.96\times10^{7}\xspace}  % coord.pc1.S: median over capture
\newcommand{\CoordPCtwodSblind}{7.21\times10^{5}\xspace}  % coord.pc2d.S: median over capture
\newcommand{\CoordUSblind}{3.57\times10^{6}\xspace}  % coord.u.S: median over capture
\newcommand{\CovRegInvariance}{6.14\times10^{-9}\xspace}  % inv.cov_reg: median over capture
\newcommand{\DetArmVarCtrl}{0.00231\xspace}  % det.armvar.ctrl: median over configuration
\newcommand{\DetArmVarIOI}{0.522\xspace}  % det.armvar.ioi: median over configuration
\newcommand{\DetCtrlJTwo}{1/28\xspace}  % det.ctl.j2: count over configuration
\newcommand{\DetCtrlRobHi}{16\xspace}  % det.ctl.robhi: max over 5 robust covariances over configuration
\newcommand{\DetCtrlRobLo}{8\xspace}  % det.ctl.roblo: min over 5 robust covariances over configuration
\newcommand{\DetIOIJTwo}{1/31\xspace}  % det.ioi.j2: count over configuration
\newcommand{\DetIOIRobHi}{29\xspace}  % det.ioi.robhi: max over 5 robust covariances over configuration
\newcommand{\DetIOIRobLo}{26\xspace}  % det.ioi.roblo: min over 5 robust covariances over configuration
\newcommand{\DetInconsistent}{26/26\xspace}  % det.inconsistent: count over configuration
\newcommand{\DetInconsistentAll}{27/31\xspace}  % det.inconsistent.all: count over configuration
\newcommand{\DetPowerMin}{1\xspace}  % det.power: min over single-site rejections over configuration
\newcommand{\DonorNamedCount}{173/185\xspace}  % predictM.donor: count over capture
\newcommand{\DstarAffine}{2\xspace}  % opdeg.affine: exact (population) over operator
\newcommand{\DstarAffineDegP}{1\xspace}  % opdeg.affine.degP: exact over operator
\newcommand{\DstarGatedLinear}{\infty\xspace}  % opdeg.gated_linear: exact (population) over operator
\newcommand{\DstarLayerNorm}{5\xspace}  % opdeg.layernorm: exact (population) over operator
\newcommand{\DstarRMSNorm}{5\xspace}  % opdeg.rmsnorm: exact (population) over operator
\newcommand{\DstarRMSNormDegP}{4\xspace}  % opdeg.rmsnorm.degP: exact over operator
\newcommand{\DstarSoftmax}{\infty\xspace}  % opdeg.softmax: exact (population) over operator
\newcommand{\EdgeBlindRho}{-0.869\xspace}  % edge.rho_blind: Spearman over capture
\newcommand{\EdgeBlindStrong}{38.7\xspace}  % edge.median_sepR_strong: median over capture
\newcommand{\EdgeBlindWeak}{874\xspace}  % edge.median_sepR_weak: median over capture
\newcommand{\EdgeCosRho}{-0.896\xspace}  % edge.spearman_cos_rho: Spearman over capture
\newcommand{\EdgeCosRhoSlope}{-0.949\xspace}  % edge.rho_cos_abs_slope: Spearman over capture
\newcommand{\EdgeFrozenEscapes}{105/185\xspace}  % m.edge.frozen.n: count over capture
\newcommand{\EdgeFrozenRho}{0.142\xspace}  % m.edge.frozen.rho: Spearman over capture
\newcommand{\EdgeInteriorEscapes}{44/185\xspace}  % m.edge.interior.n: count over capture
\newcommand{\EdgeInteriorRho}{0.41\xspace}  % m.edge.interior.rho: Spearman over capture
\newcommand{\EdgeMatchedEscapes}{9/185\xspace}  % m.edge.matched.n: count over capture
\newcommand{\EdgeMatchedMedEscaped}{2.09\xspace}  % m.edge.matched.med1: median over capture
\newcommand{\EdgeMatchedMedNot}{0.125\xspace}  % m.edge.matched.med0: median over capture
\newcommand{\EdgeMatchedRho}{0.284\xspace}  % m.edge.matched.rho: Spearman over capture
\newcommand{\EdgeN}{185\xspace}  % edge.n: count over capture
\newcommand{\EdgeNStrong}{107\xspace}  % edge.nstrong: count over capture
\newcommand{\EdgeStrongCos}{0.968\xspace}  % edge.strong_median_cos: median over capture
\newcommand{\EdgeStrongGtTen}{93/107\xspace}  % edge.strong_gt10: count over capture
\newcommand{\EdgeStrongRatio}{2.29\xspace}  % edge.strong_median_gR_over_gC: median over capture
\newcommand{\EdgeStrongSepGC}{13.2\xspace}  % edge.strong_median_sep_gC: median over capture
\newcommand{\EdgeSymmetricEscapes}{22/185\xspace}  % m.edge.symmetric.n: count over capture
\newcommand{\EdgeSymmetricRho}{0.362\xspace}  % m.edge.symmetric.rho: Spearman over capture
\newcommand{\EdgeWeakCos}{0.999\xspace}  % edge.weak_median_cos: median over capture
\newcommand{\EnvRatioMin}{23.6\xspace}  % env.rmin: min over degrees 2-4 of the ratio of medians over capture
\newcommand{\EnvSDfour}{220\xspace}  % env.s4: median over capture
\newcommand{\EnvSDtwo}{6.68\xspace}  % env.s2: median over capture
\newcommand{\EucRegInvariance}{0.867\xspace}  % inv.euc_reg: median over capture
\newcommand{\FSGPTEigengapBottomRel}{0.00295\xspace}  % fs.GPT-2.gap_bottom: median over capture
\newcommand{\FSGPTRateSlope}{-0.3\xspace}  % fs.GPT-2.rate: median over capture
\newcommand{\FSQwenEigengapBottomRel}{0.00317\xspace}  % fs.Qwen3.5-9B.gap_bottom: median over capture
\newcommand{\FSQwenRateSlope}{-0.486\xspace}  % fs.Qwen3.5-9B.rate: median over capture
\newcommand{\FigOneBCaptureMedian}{1.86\xspace}  % audit.fig1b.median: median over capture
\newcommand{\GrpCtrlEdge}{0.00162\xspace}  % grp.control.edge: median over configuration
\newcommand{\GrpCtrlInfoRatio}{1.23\times10^{11}\xspace}  % grp.control.info: median over configuration
\newcommand{\GrpIOIEdge}{0.0702\xspace}  % grp.ioi.edge: median over configuration
\newcommand{\GrpIOIInfoRatio}{2.68\times10^{4}\xspace}  % grp.ioi.info: median over configuration
\newcommand{\GrpIndEdge}{0.0112\xspace}  % grp.induction.edge: median over configuration
\newcommand{\GrpIndInfoRatio}{1.73\times10^{7}\xspace}  % grp.induction.info: median over configuration
\newcommand{\GrpIndN}{18\xspace}  % grp.induction.n: count over configuration
\newcommand{\HeadEdgeExactP}{0.0143\xspace}  % head.edge.exactp: exact one-sided Mann-Whitney p over source head
\newcommand{\HeadRho}{-1\xspace}  % head.rho: Spearman over head
\newcommand{\HeadRhoN}{10\xspace}  % head.rho.n: count over head
\newcommand{\HeadRhoP}{2.76\times10^{-7}\xspace}  % head.rho.p: exact permutation p (one-sided) over head
\newcommand{\HeadRhoPTwo}{5.51\times10^{-7}\xspace}  % head.rho.p2: exact permutation p (two-sided) over head
\newcommand{\HeldoutBeatsNull}{180/185\xspace}  % predictM.beatnull: count over capture
\newcommand{\HeldoutNullHi}{11.4\xspace}  % predictM.nullHi: max over cells over cell
\newcommand{\HeldoutNullLo}{7.02\xspace}  % predictM.nullLo: min over cells over cell
\newcommand{\HeldoutPromptBootLo}{87.4\xspace}  % predictM.promptBootLo: percent over prompt
\newcommand{\HeldoutPromptStrict}{28/38\xspace}  % predictM.promptStrict: count over (cell, prompt) group
\newcommand{\HeldoutPrompts}{38\xspace}  % predictM.prompts: count over (cell, prompt) group
\newcommand{\IdGPTAgrFlipJ}{0.6\xspace}  % id.GPT-2 agreement.flipj: median over capture
\newcommand{\IdGPTAgrFlipS}{0.125\xspace}  % id.GPT-2 agreement.flip: median over capture
\newcommand{\IdGPTAgrInfoJ}{0.649\xspace}  % id.GPT-2 agreement.infoJ: median over capture
\newcommand{\IdGPTAgrInfoS}{1.20\times10^{-4}\xspace}  % id.GPT-2 agreement.infoS: median over capture
\newcommand{\IdGPTAgrSeRatio}{32.4\xspace}  % id.GPT-2 agreement.se: median over capture
\newcommand{\IdGPTCapFlipJ}{0.55\xspace}  % id.GPT-2 capital.flipj: median over capture
\newcommand{\IdGPTCapFlipS}{0.025\xspace}  % id.GPT-2 capital.flip: median over capture
\newcommand{\IdGPTCapInfoJ}{0.51\xspace}  % id.GPT-2 capital.infoJ: median over capture
\newcommand{\IdGPTCapInfoS}{2.01\times10^{-7}\xspace}  % id.GPT-2 capital.infoS: median over capture
\newcommand{\IdGPTCapSeRatio}{618\xspace}  % id.GPT-2 capital.se: median over capture
\newcommand{\IdGPTGtFlipJ}{0.3\xspace}  % id.GPT-2 greater.flipj: median over capture
\newcommand{\IdGPTGtFlipS}{0.325\xspace}  % id.GPT-2 greater.flip: median over capture
\newcommand{\IdGPTGtInfoJ}{0.571\xspace}  % id.GPT-2 greater.infoJ: median over capture
\newcommand{\IdGPTGtInfoS}{6.69\times10^{-6}\xspace}  % id.GPT-2 greater.infoS: median over capture
\newcommand{\IdGPTGtSeRatio}{126\xspace}  % id.GPT-2 greater.se: median over capture
\newcommand{\IdGPTIOIFlipJ}{0.5\xspace}  % id.GPT-2 IOI.flipj: median over capture
\newcommand{\IdGPTIOIFlipS}{0.237\xspace}  % id.GPT-2 IOI.flip: median over capture
\newcommand{\IdGPTIOIInfoJ}{0.443\xspace}  % id.GPT-2 IOI.infoJ: median over capture
\newcommand{\IdGPTIOIInfoS}{2.95\times10^{-4}\xspace}  % id.GPT-2 IOI.infoS: median over capture
\newcommand{\IdGPTIOISeRatio}{19.3\xspace}  % id.GPT-2 IOI.se: median over capture
\newcommand{\IdGPTIndFlipJ}{0.25\xspace}  % id.GPT-2 induction.flipj: median over capture
\newcommand{\IdGPTIndFlipS}{0.075\xspace}  % id.GPT-2 induction.flip: median over capture
\newcommand{\IdGPTIndInfoJ}{0.647\xspace}  % id.GPT-2 induction.infoJ: median over capture
\newcommand{\IdGPTIndInfoS}{0.00457\xspace}  % id.GPT-2 induction.infoS: median over capture
\newcommand{\IdGPTIndSeRatio}{4.43\xspace}  % id.GPT-2 induction.se: median over capture
\newcommand{\IdQwenCapFlipJ}{0.4\xspace}  % id.Qwen3.5 capital.flipj: median over capture
\newcommand{\IdQwenCapFlipS}{0.0125\xspace}  % id.Qwen3.5 capital.flip: median over capture
\newcommand{\IdQwenCapInfoJ}{0.526\xspace}  % id.Qwen3.5 capital.infoJ: median over capture
\newcommand{\IdQwenCapInfoS}{1.48\times10^{-7}\xspace}  % id.Qwen3.5 capital.infoS: median over capture
\newcommand{\IdQwenCapSeRatio}{833\xspace}  % id.Qwen3.5 capital.se: median over capture
\newcommand{\IdQwenIOIFlipJ}{0.3\xspace}  % id.Qwen3.5 IOI.flipj: median over capture
\newcommand{\IdQwenIOIFlipS}{0.287\xspace}  % id.Qwen3.5 IOI.flip: median over capture
\newcommand{\IdQwenIOIInfoJ}{0.416\xspace}  % id.Qwen3.5 IOI.infoJ: median over capture
\newcommand{\IdQwenIOIInfoS}{3.17\times10^{-4}\xspace}  % id.Qwen3.5 IOI.infoS: median over capture
\newcommand{\IdQwenIOISeRatio}{19.9\xspace}  % id.Qwen3.5 IOI.se: median over capture
\newcommand{\IdQwenIndFlipJ}{0.6\xspace}  % id.Qwen3.5 induction.flipj: median over capture
\newcommand{\IdQwenIndFlipS}{0.35\xspace}  % id.Qwen3.5 induction.flip: median over capture
\newcommand{\IdQwenIndInfoJ}{0.676\xspace}  % id.Qwen3.5 induction.infoJ: median over capture
\newcommand{\IdQwenIndInfoS}{0.0116\xspace}  % id.Qwen3.5 induction.infoS: median over capture
\newcommand{\IdQwenIndSeRatio}{4.61\xspace}  % id.Qwen3.5 induction.se: median over capture
\newcommand{\KBlindFive}{71.4\xspace}  % ksite.blind5: exact (population) over site count
\newcommand{\KBlindFour}{62.9\xspace}  % ksite.blind4: exact (population) over site count
\newcommand{\KBlindThree}{50.0\xspace}  % ksite.blind3: exact (population) over site count
\newcommand{\KBlindTwo}{30.0\xspace}  % ksite.blind2: exact (population) over site count
\newcommand{\KCEBound}{41\xspace}  % ksite.ce.bound: exact over configuration
\newcommand{\KCEComposedBound}{76\xspace}  % ksite.ce.cb: exact over configuration
\newcommand{\KCERankExact}{59\xspace}  % ksite.ce.xrank: exact rank over F_p, generic coefficients, edges shared across arms over configuration
\newcommand{\KCEd}{5\xspace}  % ksite.ce.d: exact over configuration
\newcommand{\KCEe}{2\xspace}  % ksite.ce.e: exact over configuration
\newcommand{\KCEk}{4\xspace}  % ksite.ce.k: exact over configuration
\newcommand{\MCLLBelowRandTwo}{12/12\xspace}  % mech.ll_below_rand.d2: count over capture
\newcommand{\MCLLDegOne}{0.13\xspace}  % mech.logitlens/d1: median over capture
\newcommand{\MCLLDegTwo}{2.01\times10^{-6}\xspace}  % mech.logitlens/d2: median over capture
\newcommand{\MCLLNull}{0.409\xspace}  % mech.logitlens.null1: median across captures of per-capture fifth percentiles over five same-family resamples over capture
\newcommand{\MCLLNullTwo}{0.342\xspace}  % mech.logitlens.null2: median across captures of per-capture fifth percentiles over five same-family resamples over capture
\newcommand{\MCNullRatioOneHi}{3.15\xspace}  % mech.nullratio.d1.hi: max over the four representations of (capture-median null / capture-median share) over representation
\newcommand{\MCNullRatioOneLo}{2.05\xspace}  % mech.nullratio.d1.lo: min over the four representations of (capture-median null / capture-median share) over representation
\newcommand{\MCNullRatioTwoHi}{1.70\times10^{5}\xspace}  % mech.nullratio.d2.hi: max over the four representations of (capture-median null / capture-median share) over representation
\newcommand{\MCNullRatioTwoLo}{4.50\times10^{3}\xspace}  % mech.nullratio.d2.lo: min over the four representations of (capture-median null / capture-median share) over representation
\newcommand{\MCRandDegOne}{0.185\xspace}  % mech.randproj/d1: median over capture
\newcommand{\MCRandDegTwo}{6.25\times10^{-5}\xspace}  % mech.randproj/d2: median over capture
\newcommand{\MCRandNull}{0.379\xspace}  % mech.randproj.null1: median across captures of per-capture fifth percentiles over five same-family resamples over capture
\newcommand{\MCRandNullTwo}{0.281\xspace}  % mech.randproj.null2: median across captures of per-capture fifth percentiles over five same-family resamples over capture
\newcommand{\MCRawBelowRandTwo}{12/12\xspace}  % mech.raw_below_rand.d2: count over capture
\newcommand{\MCRawDegOne}{0.143\xspace}  % mech.raw/d1: median over capture
\newcommand{\MCRawDegTwo}{5.27\times10^{-6}\xspace}  % mech.raw/d2: median over capture
\newcommand{\MCRawNull}{0.378\xspace}  % mech.raw.null1: median across captures of per-capture fifth percentiles over five same-family resamples over capture
\newcommand{\MCRawNullTwo}{0.24\xspace}  % mech.raw.null2: median across captures of per-capture fifth percentiles over five same-family resamples over capture
\newcommand{\MCTLBelowRandTwo}{10/12\xspace}  % mech.tl_below_rand.d2: count over capture
\newcommand{\MCTLDegOne}{0.164\xspace}  % mech.tunedlens/d1: median over capture
\newcommand{\MCTLDegTwo}{2.71\times10^{-5}\xspace}  % mech.tunedlens/d2: median over capture
\newcommand{\MCTLNull}{0.362\xspace}  % mech.tunedlens.null1: median across captures of per-capture fifth percentiles over five same-family resamples over capture
\newcommand{\MCTLNullTwo}{0.212\xspace}  % mech.tunedlens.null2: median across captures of per-capture fifth percentiles over five same-family resamples over capture
\newcommand{\MGPTAgrFrozenJblind}{5/9\xspace}  % m.GPT-2 agreement.frozen.J: count over capture
\newcommand{\MGPTAgrFrozenQJ}{4.52\times10^{-6}\xspace}  % m.GPT-2 agreement.frozen.qJ: Holm-adjusted binomial p over cell
\newcommand{\MGPTAgrFrozenQS}{2.69\times10^{-12}\xspace}  % m.GPT-2 agreement.frozen.qS: Holm-adjusted binomial p over cell
\newcommand{\MGPTAgrFrozenSblind}{8/9\xspace}  % m.GPT-2 agreement.frozen.S: count over capture
\newcommand{\MGPTAgrInteriorJblind}{1/9\xspace}  % m.GPT-2 agreement.interior.J: count over capture
\newcommand{\MGPTAgrInteriorQJ}{1\xspace}  % m.GPT-2 agreement.interior.qJ: Holm-adjusted binomial p over cell
\newcommand{\MGPTAgrInteriorQS}{7.63\times10^{-15}\xspace}  % m.GPT-2 agreement.interior.qS: Holm-adjusted binomial p over cell
\newcommand{\MGPTAgrInteriorSblind}{9/9\xspace}  % m.GPT-2 agreement.interior.S: count over capture
\newcommand{\MGPTAgrMatchedJblind}{1/9\xspace}  % m.GPT-2 agreement.matched.J: count over capture
\newcommand{\MGPTAgrMatchedQJ}{1\xspace}  % m.GPT-2 agreement.matched.qJ: Holm-adjusted binomial p over cell
\newcommand{\MGPTAgrMatchedQS}{2.69\times10^{-12}\xspace}  % m.GPT-2 agreement.matched.qS: Holm-adjusted binomial p over cell
\newcommand{\MGPTAgrMatchedSblind}{8/9\xspace}  % m.GPT-2 agreement.matched.S: count over capture
\newcommand{\MGPTAgrSymmetricJblind}{1/9\xspace}  % m.GPT-2 agreement.symmetric.J: count over capture
\newcommand{\MGPTAgrSymmetricQJ}{1\xspace}  % m.GPT-2 agreement.symmetric.qJ: Holm-adjusted binomial p over cell
\newcommand{\MGPTAgrSymmetricQS}{5.37\times10^{-12}\xspace}  % m.GPT-2 agreement.symmetric.qS: Holm-adjusted binomial p over cell
\newcommand{\MGPTAgrSymmetricSblind}{8/9\xspace}  % m.GPT-2 agreement.symmetric.S: count over capture
\newcommand{\MGPTCapFrozenJblind}{15/19\xspace}  % m.GPT-2 capital.frozen.J: count over capture
\newcommand{\MGPTCapFrozenQJ}{2.63\times10^{-20}\xspace}  % m.GPT-2 capital.frozen.qJ: Holm-adjusted binomial p over cell
\newcommand{\MGPTCapFrozenQS}{2.18\times10^{-30}\xspace}  % m.GPT-2 capital.frozen.qS: Holm-adjusted binomial p over cell
\newcommand{\MGPTCapFrozenSblind}{19/19\xspace}  % m.GPT-2 capital.frozen.S: count over capture
\newcommand{\MGPTCapInteriorJblind}{0/19\xspace}  % m.GPT-2 capital.interior.J: count over capture
\newcommand{\MGPTCapInteriorQJ}{1\xspace}  % m.GPT-2 capital.interior.qJ: Holm-adjusted binomial p over cell
\newcommand{\MGPTCapInteriorQS}{2.18\times10^{-30}\xspace}  % m.GPT-2 capital.interior.qS: Holm-adjusted binomial p over cell
\newcommand{\MGPTCapInteriorSblind}{19/19\xspace}  % m.GPT-2 capital.interior.S: count over capture
\newcommand{\MGPTCapMatchedJblind}{0/19\xspace}  % m.GPT-2 capital.matched.J: count over capture
\newcommand{\MGPTCapMatchedQJ}{1\xspace}  % m.GPT-2 capital.matched.qJ: Holm-adjusted binomial p over cell
\newcommand{\MGPTCapMatchedQS}{2.18\times10^{-30}\xspace}  % m.GPT-2 capital.matched.qS: Holm-adjusted binomial p over cell
\newcommand{\MGPTCapMatchedSblind}{19/19\xspace}  % m.GPT-2 capital.matched.S: count over capture
\newcommand{\MGPTCapSymmetricJblind}{0/19\xspace}  % m.GPT-2 capital.symmetric.J: count over capture
\newcommand{\MGPTCapSymmetricQJ}{1\xspace}  % m.GPT-2 capital.symmetric.qJ: Holm-adjusted binomial p over cell
\newcommand{\MGPTCapSymmetricQS}{2.55\times10^{-30}\xspace}  % m.GPT-2 capital.symmetric.qS: Holm-adjusted binomial p over cell
\newcommand{\MGPTCapSymmetricSblind}{19/19\xspace}  % m.GPT-2 capital.symmetric.S: count over capture
\newcommand{\MGPTGtFrozenJblind}{3/20\xspace}  % m.GPT-2 greater.frozen.J: count over capture
\newcommand{\MGPTGtFrozenQJ}{0.0259\xspace}  % m.GPT-2 greater.frozen.qJ: Holm-adjusted binomial p over cell
\newcommand{\MGPTGtFrozenQS}{7.28\times10^{-32}\xspace}  % m.GPT-2 greater.frozen.qS: Holm-adjusted binomial p over cell
\newcommand{\MGPTGtFrozenSblind}{20/20\xspace}  % m.GPT-2 greater.frozen.S: count over capture
\newcommand{\MGPTGtInteriorJblind}{0/20\xspace}  % m.GPT-2 greater.interior.J: count over capture
\newcommand{\MGPTGtInteriorQJ}{1\xspace}  % m.GPT-2 greater.interior.qJ: Holm-adjusted binomial p over cell
\newcommand{\MGPTGtInteriorQS}{7.28\times10^{-32}\xspace}  % m.GPT-2 greater.interior.qS: Holm-adjusted binomial p over cell
\newcommand{\MGPTGtInteriorSblind}{20/20\xspace}  % m.GPT-2 greater.interior.S: count over capture
\newcommand{\MGPTGtMatchedJblind}{0/20\xspace}  % m.GPT-2 greater.matched.J: count over capture
\newcommand{\MGPTGtMatchedQJ}{1\xspace}  % m.GPT-2 greater.matched.qJ: Holm-adjusted binomial p over cell
\newcommand{\MGPTGtMatchedQS}{7.28\times10^{-32}\xspace}  % m.GPT-2 greater.matched.qS: Holm-adjusted binomial p over cell
\newcommand{\MGPTGtMatchedSblind}{20/20\xspace}  % m.GPT-2 greater.matched.S: count over capture
\newcommand{\MGPTGtSymmetricJblind}{1/20\xspace}  % m.GPT-2 greater.symmetric.J: count over capture
\newcommand{\MGPTGtSymmetricQJ}{1\xspace}  % m.GPT-2 greater.symmetric.qJ: Holm-adjusted binomial p over cell
\newcommand{\MGPTGtSymmetricQS}{7.28\times10^{-32}\xspace}  % m.GPT-2 greater.symmetric.qS: Holm-adjusted binomial p over cell
\newcommand{\MGPTGtSymmetricSblind}{20/20\xspace}  % m.GPT-2 greater.symmetric.S: count over capture
\newcommand{\MGPTIOIFrozenJblind}{6/34\xspace}  % m.GPT-2 ioi.frozen.J: count over capture
\newcommand{\MGPTIOIFrozenQJ}{5.39\times10^{-4}\xspace}  % m.GPT-2 ioi.frozen.qJ: Holm-adjusted binomial p over cell
\newcommand{\MGPTIOIFrozenQS}{3.39\times10^{-54}\xspace}  % m.GPT-2 ioi.frozen.qS: Holm-adjusted binomial p over cell
\newcommand{\MGPTIOIFrozenSblind}{34/34\xspace}  % m.GPT-2 ioi.frozen.S: count over capture
\newcommand{\MGPTIOIInteriorJblind}{1/34\xspace}  % m.GPT-2 ioi.interior.J: count over capture
\newcommand{\MGPTIOIInteriorQJ}{1\xspace}  % m.GPT-2 ioi.interior.qJ: Holm-adjusted binomial p over cell
\newcommand{\MGPTIOIInteriorQS}{3.05\times10^{-54}\xspace}  % m.GPT-2 ioi.interior.qS: Holm-adjusted binomial p over cell
\newcommand{\MGPTIOIInteriorSblind}{34/34\xspace}  % m.GPT-2 ioi.interior.S: count over capture
\newcommand{\MGPTIOIMatchedJblind}{2/34\xspace}  % m.GPT-2 ioi.matched.J: count over capture
\newcommand{\MGPTIOIMatchedQJ}{1\xspace}  % m.GPT-2 ioi.matched.qJ: Holm-adjusted binomial p over cell
\newcommand{\MGPTIOIMatchedQS}{3.39\times10^{-54}\xspace}  % m.GPT-2 ioi.matched.qS: Holm-adjusted binomial p over cell
\newcommand{\MGPTIOIMatchedSblind}{34/34\xspace}  % m.GPT-2 ioi.matched.S: count over capture
\newcommand{\MGPTIOISymmetricJblind}{1/34\xspace}  % m.GPT-2 ioi.symmetric.J: count over capture
\newcommand{\MGPTIOISymmetricQJ}{1\xspace}  % m.GPT-2 ioi.symmetric.qJ: Holm-adjusted binomial p over cell
\newcommand{\MGPTIOISymmetricQS}{2.61\times10^{-48}\xspace}  % m.GPT-2 ioi.symmetric.qS: Holm-adjusted binomial p over cell
\newcommand{\MGPTIOISymmetricSblind}{32/34\xspace}  % m.GPT-2 ioi.symmetric.S: count over capture
\newcommand{\MGPTIndFrozenJblind}{35/35\xspace}  % m.GPT-2 induction.frozen.J: count over capture
\newcommand{\MGPTIndFrozenQJ}{8.47\times10^{-56}\xspace}  % m.GPT-2 induction.frozen.qJ: Holm-adjusted binomial p over cell
\newcommand{\MGPTIndFrozenQS}{1.04\times10^{-52}\xspace}  % m.GPT-2 induction.frozen.qS: Holm-adjusted binomial p over cell
\newcommand{\MGPTIndFrozenSblind}{34/35\xspace}  % m.GPT-2 induction.frozen.S: count over capture
\newcommand{\MGPTIndInteriorJblind}{33/35\xspace}  % m.GPT-2 induction.interior.J: count over capture
\newcommand{\MGPTIndInteriorQJ}{7.68\times10^{-50}\xspace}  % m.GPT-2 induction.interior.qJ: Holm-adjusted binomial p over cell
\newcommand{\MGPTIndInteriorQS}{8.47\times10^{-56}\xspace}  % m.GPT-2 induction.interior.qS: Holm-adjusted binomial p over cell
\newcommand{\MGPTIndInteriorSblind}{35/35\xspace}  % m.GPT-2 induction.interior.S: count over capture
\newcommand{\MGPTIndMatchedJblind}{3/35\xspace}  % m.GPT-2 induction.matched.J: count over capture
\newcommand{\MGPTIndMatchedQJ}{0.509\xspace}  % m.GPT-2 induction.matched.qJ: Holm-adjusted binomial p over cell
\newcommand{\MGPTIndMatchedQS}{6.91\times10^{-50}\xspace}  % m.GPT-2 induction.matched.qS: Holm-adjusted binomial p over cell
\newcommand{\MGPTIndMatchedSblind}{33/35\xspace}  % m.GPT-2 induction.matched.S: count over capture
\newcommand{\MGPTIndSymmetricJblind}{13/35\xspace}  % m.GPT-2 induction.symmetric.J: count over capture
\newcommand{\MGPTIndSymmetricQJ}{1.31\times10^{-11}\xspace}  % m.GPT-2 induction.symmetric.qJ: Holm-adjusted binomial p over cell
\newcommand{\MGPTIndSymmetricQS}{7.68\times10^{-50}\xspace}  % m.GPT-2 induction.symmetric.qS: Holm-adjusted binomial p over cell
\newcommand{\MGPTIndSymmetricSblind}{33/35\xspace}  % m.GPT-2 induction.symmetric.S: count over capture
\newcommand{\MQIOIFrozenJblind}{7/15\xspace}  % m.Qwen3-8B ioi.frozen.J: count over capture
\newcommand{\MQIOIFrozenQJ}{1.65\times10^{-7}\xspace}  % m.Qwen3-8B ioi.frozen.qJ: Holm-adjusted binomial p over cell
\newcommand{\MQIOIFrozenQS}{2.73\times10^{-21}\xspace}  % m.Qwen3-8B ioi.frozen.qS: Holm-adjusted binomial p over cell
\newcommand{\MQIOIFrozenSblind}{14/15\xspace}  % m.Qwen3-8B ioi.frozen.S: count over capture
\newcommand{\MQIOIInteriorJblind}{1/15\xspace}  % m.Qwen3-8B ioi.interior.J: count over capture
\newcommand{\MQIOIInteriorQJ}{1\xspace}  % m.Qwen3-8B ioi.interior.qJ: Holm-adjusted binomial p over cell
\newcommand{\MQIOIInteriorQS}{2.18\times10^{-21}\xspace}  % m.Qwen3-8B ioi.interior.qS: Holm-adjusted binomial p over cell
\newcommand{\MQIOIInteriorSblind}{14/15\xspace}  % m.Qwen3-8B ioi.interior.S: count over capture
\newcommand{\MQIOIMatchedJblind}{0/15\xspace}  % m.Qwen3-8B ioi.matched.J: count over capture
\newcommand{\MQIOIMatchedQJ}{1\xspace}  % m.Qwen3-8B ioi.matched.qJ: Holm-adjusted binomial p over cell
\newcommand{\MQIOIMatchedQS}{7.46\times10^{-19}\xspace}  % m.Qwen3-8B ioi.matched.qS: Holm-adjusted binomial p over cell
\newcommand{\MQIOIMatchedSblind}{13/15\xspace}  % m.Qwen3-8B ioi.matched.S: count over capture
\newcommand{\MQIOISymmetricJblind}{0/15\xspace}  % m.Qwen3-8B ioi.symmetric.J: count over capture
\newcommand{\MQIOISymmetricQJ}{1\xspace}  % m.Qwen3-8B ioi.symmetric.qJ: Holm-adjusted binomial p over cell
\newcommand{\MQIOISymmetricQS}{1.28\times10^{-12}\xspace}  % m.Qwen3-8B ioi.symmetric.qS: Holm-adjusted binomial p over cell
\newcommand{\MQIOISymmetricSblind}{10/15\xspace}  % m.Qwen3-8B ioi.symmetric.S: count over capture
\newcommand{\MQIndFrozenJblind}{1/11\xspace}  % m.Qwen3-8B induction.frozen.J: count over capture
\newcommand{\MQIndFrozenQJ}{0.243\xspace}  % m.Qwen3-8B induction.frozen.qJ: Holm-adjusted binomial p over cell
\newcommand{\MQIndFrozenQS}{3.08\times10^{-15}\xspace}  % m.Qwen3-8B induction.frozen.qS: Holm-adjusted binomial p over cell
\newcommand{\MQIndFrozenSblind}{10/11\xspace}  % m.Qwen3-8B induction.frozen.S: count over capture
\newcommand{\MQIndInteriorJblind}{0/11\xspace}  % m.Qwen3-8B induction.interior.J: count over capture
\newcommand{\MQIndInteriorQJ}{1\xspace}  % m.Qwen3-8B induction.interior.qJ: Holm-adjusted binomial p over cell
\newcommand{\MQIndInteriorQS}{7.15\times10^{-18}\xspace}  % m.Qwen3-8B induction.interior.qS: Holm-adjusted binomial p over cell
\newcommand{\MQIndInteriorSblind}{11/11\xspace}  % m.Qwen3-8B induction.interior.S: count over capture
\newcommand{\MQIndMatchedJblind}{0/11\xspace}  % m.Qwen3-8B induction.matched.J: count over capture
\newcommand{\MQIndMatchedQJ}{1\xspace}  % m.Qwen3-8B induction.matched.qJ: Holm-adjusted binomial p over cell
\newcommand{\MQIndMatchedQS}{3.08\times10^{-15}\xspace}  % m.Qwen3-8B induction.matched.qS: Holm-adjusted binomial p over cell
\newcommand{\MQIndMatchedSblind}{10/11\xspace}  % m.Qwen3-8B induction.matched.S: count over capture
\newcommand{\MQIndSymmetricJblind}{0/11\xspace}  % m.Qwen3-8B induction.symmetric.J: count over capture
\newcommand{\MQIndSymmetricQJ}{1\xspace}  % m.Qwen3-8B induction.symmetric.qJ: Holm-adjusted binomial p over cell
\newcommand{\MQIndSymmetricQS}{7.06\times10^{-11}\xspace}  % m.Qwen3-8B induction.symmetric.qS: Holm-adjusted binomial p over cell
\newcommand{\MQIndSymmetricSblind}{8/11\xspace}  % m.Qwen3-8B induction.symmetric.S: count over capture
\newcommand{\MQwenCapFrozenJblind}{6/6\xspace}  % m.Qwen3.5 capital.frozen.J: count over capture
\newcommand{\MQwenCapFrozenQJ}{1.46\times10^{-9}\xspace}  % m.Qwen3.5 capital.frozen.qJ: Holm-adjusted binomial p over cell
\newcommand{\MQwenCapFrozenQS}{2.44\times10^{-10}\xspace}  % m.Qwen3.5 capital.frozen.qS: Holm-adjusted binomial p over cell
\newcommand{\MQwenCapFrozenSblind}{6/6\xspace}  % m.Qwen3.5 capital.frozen.S: count over capture
\newcommand{\MQwenCapInteriorJblind}{0/6\xspace}  % m.Qwen3.5 capital.interior.J: count over capture
\newcommand{\MQwenCapInteriorQJ}{1\xspace}  % m.Qwen3.5 capital.interior.qJ: Holm-adjusted binomial p over cell
\newcommand{\MQwenCapInteriorQS}{2.44\times10^{-10}\xspace}  % m.Qwen3.5 capital.interior.qS: Holm-adjusted binomial p over cell
\newcommand{\MQwenCapInteriorSblind}{6/6\xspace}  % m.Qwen3.5 capital.interior.S: count over capture
\newcommand{\MQwenCapMatchedJblind}{0/6\xspace}  % m.Qwen3.5 capital.matched.J: count over capture
\newcommand{\MQwenCapMatchedQJ}{1\xspace}  % m.Qwen3.5 capital.matched.qJ: Holm-adjusted binomial p over cell
\newcommand{\MQwenCapMatchedQS}{2.44\times10^{-10}\xspace}  % m.Qwen3.5 capital.matched.qS: Holm-adjusted binomial p over cell
\newcommand{\MQwenCapMatchedSblind}{6/6\xspace}  % m.Qwen3.5 capital.matched.S: count over capture
\newcommand{\MQwenCapSymmetricJblind}{1/6\xspace}  % m.Qwen3.5 capital.symmetric.J: count over capture
\newcommand{\MQwenCapSymmetricQJ}{0.987\xspace}  % m.Qwen3.5 capital.symmetric.qJ: Holm-adjusted binomial p over cell
\newcommand{\MQwenCapSymmetricQS}{4.88\times10^{-10}\xspace}  % m.Qwen3.5 capital.symmetric.qS: Holm-adjusted binomial p over cell
\newcommand{\MQwenCapSymmetricSblind}{6/6\xspace}  % m.Qwen3.5 capital.symmetric.S: count over capture
\newcommand{\MQwenIOIFrozenJblind}{16/20\xspace}  % m.Qwen3.5 ioi.frozen.J: count over capture
\newcommand{\MQwenIOIFrozenQJ}{9.23\times10^{-22}\xspace}  % m.Qwen3.5 ioi.frozen.qJ: Holm-adjusted binomial p over cell
\newcommand{\MQwenIOIFrozenQS}{7.28\times10^{-32}\xspace}  % m.Qwen3.5 ioi.frozen.qS: Holm-adjusted binomial p over cell
\newcommand{\MQwenIOIFrozenSblind}{20/20\xspace}  % m.Qwen3.5 ioi.frozen.S: count over capture
\newcommand{\MQwenIOIInteriorJblind}{1/20\xspace}  % m.Qwen3.5 ioi.interior.J: count over capture
\newcommand{\MQwenIOIInteriorQJ}{1\xspace}  % m.Qwen3.5 ioi.interior.qJ: Holm-adjusted binomial p over cell
\newcommand{\MQwenIOIInteriorQS}{7.28\times10^{-32}\xspace}  % m.Qwen3.5 ioi.interior.qS: Holm-adjusted binomial p over cell
\newcommand{\MQwenIOIInteriorSblind}{20/20\xspace}  % m.Qwen3.5 ioi.interior.S: count over capture
\newcommand{\MQwenIOIMatchedJblind}{3/20\xspace}  % m.Qwen3.5 ioi.matched.J: count over capture
\newcommand{\MQwenIOIMatchedQJ}{0.13\xspace}  % m.Qwen3.5 ioi.matched.qJ: Holm-adjusted binomial p over cell
\newcommand{\MQwenIOIMatchedQS}{7.28\times10^{-32}\xspace}  % m.Qwen3.5 ioi.matched.qS: Holm-adjusted binomial p over cell
\newcommand{\MQwenIOIMatchedSblind}{20/20\xspace}  % m.Qwen3.5 ioi.matched.S: count over capture
\newcommand{\MQwenIOISymmetricJblind}{3/20\xspace}  % m.Qwen3.5 ioi.symmetric.J: count over capture
\newcommand{\MQwenIOISymmetricQJ}{0.117\xspace}  % m.Qwen3.5 ioi.symmetric.qJ: Holm-adjusted binomial p over cell
\newcommand{\MQwenIOISymmetricQS}{1.58\times10^{-26}\xspace}  % m.Qwen3.5 ioi.symmetric.qS: Holm-adjusted binomial p over cell
\newcommand{\MQwenIOISymmetricSblind}{18/20\xspace}  % m.Qwen3.5 ioi.symmetric.S: count over capture
\newcommand{\MQwenIndFrozenJblind}{11/16\xspace}  % m.Qwen3.5 induction.frozen.J: count over capture
\newcommand{\MQwenIndFrozenQJ}{6.49\times10^{-14}\xspace}  % m.Qwen3.5 induction.frozen.qJ: Holm-adjusted binomial p over cell
\newcommand{\MQwenIndFrozenQS}{1.71\times10^{-20}\xspace}  % m.Qwen3.5 induction.frozen.qS: Holm-adjusted binomial p over cell
\newcommand{\MQwenIndFrozenSblind}{14/16\xspace}  % m.Qwen3.5 induction.frozen.S: count over capture
\newcommand{\MQwenIndInteriorJblind}{7/16\xspace}  % m.Qwen3.5 induction.interior.J: count over capture
\newcommand{\MQwenIndInteriorQJ}{5.15\times10^{-7}\xspace}  % m.Qwen3.5 induction.interior.qJ: Holm-adjusted binomial p over cell
\newcommand{\MQwenIndInteriorQS}{7.28\times10^{-23}\xspace}  % m.Qwen3.5 induction.interior.qS: Holm-adjusted binomial p over cell
\newcommand{\MQwenIndInteriorSblind}{15/16\xspace}  % m.Qwen3.5 induction.interior.S: count over capture
\newcommand{\MQwenIndMatchedJblind}{0/16\xspace}  % m.Qwen3.5 induction.matched.J: count over capture
\newcommand{\MQwenIndMatchedQJ}{1\xspace}  % m.Qwen3.5 induction.matched.qJ: Holm-adjusted binomial p over cell
\newcommand{\MQwenIndMatchedQS}{3.11\times10^{-18}\xspace}  % m.Qwen3.5 induction.matched.qS: Holm-adjusted binomial p over cell
\newcommand{\MQwenIndMatchedSblind}{13/16\xspace}  % m.Qwen3.5 induction.matched.S: count over capture
\newcommand{\MQwenIndSymmetricJblind}{2/16\xspace}  % m.Qwen3.5 induction.symmetric.J: count over capture
\newcommand{\MQwenIndSymmetricQJ}{0.476\xspace}  % m.Qwen3.5 induction.symmetric.qJ: Holm-adjusted binomial p over cell
\newcommand{\MQwenIndSymmetricQS}{1.64\times10^{-9}\xspace}  % m.Qwen3.5 induction.symmetric.qS: Holm-adjusted binomial p over cell
\newcommand{\MQwenIndSymmetricSblind}{8/16\xspace}  % m.Qwen3.5 induction.symmetric.S: count over capture
\newcommand{\MatchedN}{185\xspace}  % predictM.n: count over capture
\newcommand{\MfJThree}{0.043\xspace}  % mf.j3: median over capture
\newcommand{\MfJWorse}{22/31\xspace}  % mf.jworse: count over capture
\newcommand{\MfMargin}{0.10\xspace}  % mf.margin: declared over -
\newcommand{\MfMaxAbsMax}{0.30\xspace}  % mf.maxabs: max over capture x row
\newcommand{\MfSThree}{7.45\xspace}  % mf.s3: median over capture
\newcommand{\MfSWorse}{31/31\xspace}  % mf.sworse: count over capture
\newcommand{\MfSwingMed}{1.65\xspace}  % mf.swing.med: median over captures over capture
\newcommand{\MfSwingMin}{0.67\xspace}  % mf.swing.min: min over captures over capture
\newcommand{\MfSwingThr}{0.5\xspace}  % mf.swing.thr: declared over -
\newcommand{\MfZero}{0.010\xspace}  % mf.zero: median over capture
\newcommand{\MisJ}{0.764\xspace}  % mis.J: median over capture
\newcommand{\MisN}{31\xspace}  % mis.n: count over capture
\newcommand{\MisS}{31.1\xspace}  % mis.S: median over capture
\newcommand{\NAlpha}{7\xspace}  % donor.nalpha: count over grid point
\newcommand{\OfatBoth}{143/185\xspace}  % ofat.row.both: count over capture
\newcommand{\OfatDonorBoth}{149/185\xspace}  % ofat.donor.both: count over capture
\newcommand{\OfatDonorCount}{149/185\xspace}  % ofat.donor: count over capture
\newcommand{\OfatDonorNeither}{12/185\xspace}  % ofat.donor.neither: count over capture
\newcommand{\OfatDonorOnlyClampedCount}{0/185\xspace}  % ofat.donor.onlyC: count over capture
\newcommand{\OfatDonorOnlyRecompute}{24/185\xspace}  % ofat.donor.onlyR: count over capture
\newcommand{\OfatNeither}{13/185\xspace}  % ofat.row.neither: count over capture
\newcommand{\OfatNetDifference}{27\xspace}  % ofat.row.net: difference of counts over capture
\newcommand{\OfatOnlyClampedCount}{1/185\xspace}  % ofat.row.onlyC: count over capture
\newcommand{\OfatOnlyRecompute}{28/185\xspace}  % ofat.row.onlyR: count over capture
\newcommand{\OfatPromptStrict}{17/38\xspace}  % ofat.prompt: count over (cell, prompt) group
\newcommand{\OfatRowCount}{144/185\xspace}  % ofat.row: count over capture
\newcommand{\OfatSpearman}{0.587\xspace}  % ofat.rho: Spearman over capture
\newcommand{\PCRandPOne}{0.439\xspace}  % pc.rndp1: min over noise levels over noise level
\newcommand{\PCResidWorst}{0.0185\xspace}  % pc.worst: max over noise level
\newcommand{\PCResidZero}{6.92\times10^{-7}\xspace}  % pc.zero: value over noise level
\newcommand{\PredAgrGPTCount}{3/3\xspace}  % predict.PredAgrGPT.n: count over capture
\newcommand{\PredAgrGPTMCount}{8/9\xspace}  % predictM.PredAgrGPT.n: count over capture
\newcommand{\PredAgrGPTMOfat}{8/9\xspace}  % ofat.PredAgrGPT: count over capture
\newcommand{\PredAgrGPTMRatio}{128\xspace}  % predictM.PredAgrGPT: median over capture
\newcommand{\PredAgrGPTRatio}{372\xspace}  % predict.PredAgrGPT: median over capture
\newcommand{\PredCapGPTCount}{11/11\xspace}  % predict.PredCapGPT.n: count over capture
\newcommand{\PredCapGPTMCount}{19/19\xspace}  % predictM.PredCapGPT.n: count over capture
\newcommand{\PredCapGPTMOfat}{19/19\xspace}  % ofat.PredCapGPT: count over capture
\newcommand{\PredCapGPTMRatio}{2516\xspace}  % predictM.PredCapGPT: median over capture
\newcommand{\PredCapGPTRatio}{6413\xspace}  % predict.PredCapGPT: median over capture
\newcommand{\PredCapQwenCount}{5/5\xspace}  % predict.PredCapQwen.n: count over capture
\newcommand{\PredCapQwenMCount}{6/6\xspace}  % predictM.PredCapQwen.n: count over capture
\newcommand{\PredCapQwenMOfat}{6/6\xspace}  % ofat.PredCapQwen: count over capture
\newcommand{\PredCapQwenMRatio}{3972\xspace}  % predictM.PredCapQwen: median over capture
\newcommand{\PredCapQwenRatio}{6768\xspace}  % predict.PredCapQwen: median over capture
\newcommand{\PredCells}{10\xspace}  % predictM.cells: count over cell
\newcommand{\PredGPTIOICount}{12/12\xspace}  % predict.PredGPTIOI.n: count over capture
\newcommand{\PredGPTIOIMCount}{34/34\xspace}  % predictM.PredGPTIOI.n: count over capture
\newcommand{\PredGPTIOIMOfat}{26/34\xspace}  % ofat.PredGPTIOI: count over capture
\newcommand{\PredGPTIOIMRatio}{244\xspace}  % predictM.PredGPTIOI: median over capture
\newcommand{\PredGPTIOIRatio}{6489\xspace}  % predict.PredGPTIOI: median over capture
\newcommand{\PredGtGPTCount}{12/12\xspace}  % predict.PredGtGPT.n: count over capture
\newcommand{\PredGtGPTMCount}{20/20\xspace}  % predictM.PredGtGPT.n: count over capture
\newcommand{\PredGtGPTMOfat}{20/20\xspace}  % ofat.PredGtGPT: count over capture
\newcommand{\PredGtGPTMRatio}{562\xspace}  % predictM.PredGtGPT: median over capture
\newcommand{\PredGtGPTRatio}{1328\xspace}  % predict.PredGtGPT: median over capture
\newcommand{\PredIndGPTCount}{11/12\xspace}  % predict.PredIndGPT.n: count over capture
\newcommand{\PredIndGPTMCount}{30/35\xspace}  % predictM.PredIndGPT.n: count over capture
\newcommand{\PredIndGPTMOfat}{20/35\xspace}  % ofat.PredIndGPT: count over capture
\newcommand{\PredIndGPTMRatio}{30.1\xspace}  % predictM.PredIndGPT: median over capture
\newcommand{\PredIndGPTRatio}{101\xspace}  % predict.PredIndGPT: median over capture
\newcommand{\PredIndQwenCount}{10/12\xspace}  % predict.PredIndQwen.n: count over capture
\newcommand{\PredIndQwenMCount}{12/16\xspace}  % predictM.PredIndQwen.n: count over capture
\newcommand{\PredIndQwenMOfat}{9/16\xspace}  % ofat.PredIndQwen: count over capture
\newcommand{\PredIndQwenMRatio}{22.4\xspace}  % predictM.PredIndQwen: median over capture
\newcommand{\PredIndQwenRatio}{63.5\xspace}  % predict.PredIndQwen: median over capture
\newcommand{\PredMTotalCount}{171/185\xspace}  % predictM.total: count over capture
\newcommand{\PredQIOIMCount}{12/15\xspace}  % predictM.PredQIOI.n: count over capture
\newcommand{\PredQIOIMOfat}{12/15\xspace}  % ofat.PredQIOI: count over capture
\newcommand{\PredQIOIMRatio}{60.0\xspace}  % predictM.PredQIOI: median over capture
\newcommand{\PredQIndMCount}{10/11\xspace}  % predictM.PredQInd.n: count over capture
\newcommand{\PredQIndMOfat}{9/11\xspace}  % ofat.PredQInd: count over capture
\newcommand{\PredQIndMRatio}{66.4\xspace}  % predictM.PredQInd: median over capture
\newcommand{\PredQwenIOICount}{15/15\xspace}  % predict.PredQwenIOI.n: count over capture
\newcommand{\PredQwenIOIMCount}{20/20\xspace}  % predictM.PredQwenIOI.n: count over capture
\newcommand{\PredQwenIOIMOfat}{15/20\xspace}  % ofat.PredQwenIOI: count over capture
\newcommand{\PredQwenIOIMRatio}{117\xspace}  % predictM.PredQwenIOI: median over capture
\newcommand{\PredQwenIOIRatio}{371\xspace}  % predict.PredQwenIOI: median over capture
\newcommand{\PredTotalCount}{79/82\xspace}  % predict.total: count over capture
\newcommand{\QCCtrlN}{28\xspace}  % qc.ctrl.n: count over configuration
\newcommand{\QCCtrlRejS}{0/28\xspace}  % qc.ctrl.rejS: count over configuration
\newcommand{\QCEdgeRTwo}{0.111\xspace}  % qc.edge_r2: median over configuration
\newcommand{\QCEdgeVsCtrl}{43.3\xspace}  % qc.ctrl.edgeratio: ratio of medians over configuration
\newcommand{\QCInfoRatio}{2.68\times10^{4}\xspace}  % qc.info_ratio: median over configuration
\newcommand{\QCInfoRatioOwnChart}{5977\xspace}  % audit.qc.own_whitening: median over configuration
\newcommand{\QCN}{31\xspace}  % qc.n: count over configuration
\newcommand{\QCRejS}{26/31\xspace}  % qc.rejS: count over configuration
\newcommand{\RCChartM}{1.07\xspace}  % rc.chart.median_abs_m: median over capture
\newcommand{\RCDropped}{15\xspace}  % rc.dropped: count over capture
\newcommand{\RCGPTIndRblind}{19/35\xspace}  % rc.GPT-2 induction.R: count over capture
\newcommand{\RCGPTIndXseedRblind}{1/28\xspace}  % rcXseed.GPT-2 induction.R: count over capture
\newcommand{\RCHoldoutPooledR}{75/170\xspace}  % rcHoldout.pooledR: count over capture
\newcommand{\RCMutual}{7/10\xspace}  % rc.mutual: count over cell
\newcommand{\RCPooledR}{103/170\xspace}  % rc.pooledR: count over capture
\newcommand{\RCUchartChartM}{0.0359\xspace}  % rcUchart.chart.median_abs_m: median over capture
\newcommand{\RCUchartMutual}{4/10\xspace}  % rcUchart.mutual: count over cell
\newcommand{\RCXseedMutual}{3/8\xspace}  % rcXseed.mutual: count over cell
\newcommand{\RCXseedPooledR}{26/120\xspace}  % rcXseed.pooledR: count over capture
\newcommand{\RepCaptures}{576\xspace}  % rep.n: count over capture
\newcommand{\RepCtlHomo}{0/268\xspace}  % rep.s8.ctl.homo: count over configuration
\newcommand{\RepCtlN}{268\xspace}  % rep.s8.ctl.n: count over configuration
\newcommand{\RepCtlResolved}{111/268\xspace}  % rep.ctl.resolved: count over capture
\newcommand{\RepCtlRmsMed}{2.33\times10^{-4}\xspace}  % rep.ctl.rms.med: median over capture
\newcommand{\RepCtlRobHi}{142\xspace}  % rep.s8.ctl.hi: max over robust estimators over configuration
\newcommand{\RepCtlRobLo}{69\xspace}  % rep.s8.ctl.lo: min over robust estimators over configuration
\newcommand{\RepErrors}{0\xspace}  % rep.errors: count over capture
\newcommand{\RepFloorMed}{2.80\times10^{-5}\xspace}  % rep.floor.med: median over capture
\newcommand{\RepFloorMedCtl}{2.79\times10^{-5}\xspace}  % rep.floor.med.ctl: median over capture
\newcommand{\RepHAbsCount}{8/8\xspace}  % rep.h1abs: count over template
\newcommand{\RepHRelCount}{8/8\xspace}  % rep.h1rel: count over template
\newcommand{\RepInconsistent}{225/281\xspace}  % rep.s8.incons: count over configuration
\newcommand{\RepPrompts}{24\xspace}  % rep.prompts: count over prompt
\newcommand{\RepPubHomo}{223/281\xspace}  % rep.s8.pub.homo: count over configuration
\newcommand{\RepPubRelMax}{6.86\xspace}  % rep.pub.rel.max: max over capture
\newcommand{\RepPubRelPninety}{1.85\xspace}  % rep.pub.rel.p90: 90th percentile over capture
\newcommand{\RepPubResolved}{281/281\xspace}  % rep.pub.resolved: count over capture
\newcommand{\RepPubRmsMed}{0.012\xspace}  % rep.pub.rms.med: median over capture
\newcommand{\RepSSevenJ}{15/281\xspace}  % rep.s7.J: count over capture
\newcommand{\RepSSevenS}{155/281\xspace}  % rep.s7.S: count over capture
\newcommand{\RepSSixWorse}{281/281\xspace}  % rep.s6: count over capture
\newcommand{\RepScreenedOut}{27\xspace}  % rep.screened: count over capture
\newcommand{\RepTemplates}{8\xspace}  % rep.templates: count over template
\newcommand{\RepVMax}{0.0181\xspace}  % rep.V.max: max over template
\newcommand{\RepVMed}{0.0128\xspace}  % rep.V.med: median over template
\newcommand{\RepVMin}{0.00673\xspace}  % rep.V.min: min over template
\newcommand{\RepVTEight}{0.0121\xspace}  % rep.T8.V: median of prompt medians over template
\newcommand{\RepVTFive}{0.0118\xspace}  % rep.T5.V: median of prompt medians over template
\newcommand{\RepVTFour}{0.0164\xspace}  % rep.T4.V: median of prompt medians over template
\newcommand{\RepVTOne}{0.0181\xspace}  % rep.T1.V: median of prompt medians over template
\newcommand{\RepVTSeven}{0.0147\xspace}  % rep.T7.V: median of prompt medians over template
\newcommand{\RepVTSix}{0.00904\xspace}  % rep.T6.V: median of prompt medians over template
\newcommand{\RepVTThree}{0.00673\xspace}  % rep.T3.V: median of prompt medians over template
\newcommand{\RepVTTwo}{0.0135\xspace}  % rep.T2.V: median of prompt medians over template
\newcommand{\RepVrelMax}{0.882\xspace}  % rep.Vrel.max: max over template
\newcommand{\RepVrelMed}{0.629\xspace}  % rep.Vrel.med: median over template
\newcommand{\RepVrelMin}{0.519\xspace}  % rep.Vrel.min: min over template
\newcommand{\RepVrelTEight}{0.589\xspace}  % rep.T8.Vrel: median of prompt medians over template
\newcommand{\RepVrelTFive}{0.661\xspace}  % rep.T5.Vrel: median of prompt medians over template
\newcommand{\RepVrelTFour}{0.537\xspace}  % rep.T4.Vrel: median of prompt medians over template
\newcommand{\RepVrelTOne}{0.853\xspace}  % rep.T1.Vrel: median of prompt medians over template
\newcommand{\RepVrelTSeven}{0.882\xspace}  % rep.T7.Vrel: median of prompt medians over template
\newcommand{\RepVrelTSix}{0.771\xspace}  % rep.T6.Vrel: median of prompt medians over template
\newcommand{\RepVrelTThree}{0.519\xspace}  % rep.T3.Vrel: median of prompt medians over template
\newcommand{\RepVrelTTwo}{0.598\xspace}  % rep.T2.Vrel: median of prompt medians over template
\newcommand{\ReparamInvariance}{4.53\times10^{-7}\xspace}  % reparam.invariance: max over cells of capture medians of the per-capture max relative eigenvalue change over capture
\newcommand{\RespAddThreeJJ}{0.60\xspace}  % resp.add3.JJ: median over 10 partitions of the capture median over capture
\newcommand{\RespAddThreeSJ}{0.70\xspace}  % resp.add3.SJ: median over 10 partitions of the capture median over capture
\newcommand{\RespFullThreeJJ}{0.42\xspace}  % resp.full3.JJ: median over 10 partitions of the capture median over capture
\newcommand{\RespSJNegMin}{28\xspace}  % resp.sjneg.min: min over 10 partitions over capture
\newcommand{\RespSJR}{-439\xspace}  % resp.sj: same over capture
\newcommand{\ReverseSurvivorsWeak}{1/9\xspace}  % edge.matched.survivors_weak: count over capture
\newcommand{\TLCalibOverlap}{113/224\xspace}  % mech.tl_calibration_overlap: count over row
\newcommand{\ThreeDegThreeDim}{20\xspace}  % three.deg3.dim: exact over class
\newcommand{\ThreeDegThreeLogRatio}{7.7\xspace}  % three.deg3.logr: median over capture
\newcommand{\ThreeDegThreeReverse}{0/12\xspace}  % three.deg3.above: count over capture
\newcommand{\ThreeDegThreeSblind}{12/12\xspace}  % three.deg3.below: count over capture
\newcommand{\ThreeDegTwoDim}{10\xspace}  % three.deg2.dim: exact over class
\newcommand{\ThreeDegTwoLogRatio}{3.87\xspace}  % three.deg2.logr: median over capture
\newcommand{\ThreeDegTwoSblind}{12/12\xspace}  % three.deg2.below: count over capture

\theoremstyle{plain}
\newtheorem{theorem}{Theorem}
\newtheorem{proposition}[theorem]{Proposition}
\newtheorem{lemma}[theorem]{Lemma}

\theoremstyle{definition}
\newtheorem{definition}[theorem]{Definition}
\newtheorem{remark}[theorem]{Remark}

\tcbset{colback=black!3,colframe=black!45,boxrule=0.5pt,arc=1pt,left=5pt,right=5pt,top=3pt,bottom=3pt,
        fonttitle=\bfseries\small,coltitle=black,colbacktitle=black!10,before skip=8pt,after skip=8pt}

\title{What Can Activation Patching Actually Establish?\\
       Reachable-State Geometry and the Limits of Mechanistic Evidence}

\author{Ankush Kadu\thanks{Corresponding author.}\\
  QpiAI, India\\
  \texttt{ankush.k@qpiai.tech} \And
  Ananth Kalyanasundaram\\
  QpiAI, India\\
  \texttt{ananth.k@qpiai.tech} \And
  Aswanth Krishnan\\
  QpiAI, India\\
  \texttt{aswanth.k@qpiai.tech}}

\begin{document}
\maketitle

\begin{abstract}
Activation patching is a central tool of mechanistic interpretability, and circuit explanations are often built from
single-site measurements. Suppose patching a head moves a logit difference by $+0.5$. Does the head still contribute
$+0.5$ after an upstream head changes? Explanations that reuse such effects assume it does, yet patching one site at a
time generally cannot test this assumption. We give a framework that determines, from the internal states an
experiment reaches and before any output is read, which candidate mechanisms it can distinguish. For idealised
connections, we prove which polynomial interactions single-site patching can miss on affine, normalisation, softmax and
gated connections; on affine connections it misses $\KBlindTwo\%$ of the dimensions of the cubic hypothesis class with
two sites and $\KBlindFive\%$ with five. In $\MatchedN$ screened runs in GPT-2 small, Qwen3-8B and Qwen3.5-9B, the interaction predicted by the connection
was over ten times more visible, in mean square, under joint patching (both sites at once) than under single-site
patching in $\PredMTotalCount$ runs, and this advantage beat that of $95\%$ of random directions in
$\HeldoutBeatsNull$. We then propose a direct test, the four-point transport contrast, which executes the two
interventions single-site patching omits. On the indirect-object-identification circuit in GPT-2 small, with the
downstream Name Mover head's output fixed, changing the upstream S-Inhibition head changed the Name Mover's effect by
only $\AdqRmsTPctEffect\%$ of its typical size in the median run (root mean square over tested states), and a
preregistered replication on $\RepPrompts$ new prompts met both size criteria in all $\RepTemplates$ templates. We also identify two pitfalls in comparing patching designs and provide a reporting checklist. Together, these
results show researchers, before any readout is scored, what a patching experiment can establish, and test directly whether a head's measured effect carries over.
\end{abstract}

\begingroup\let\thefootnote\relax\footnotetext{This paper was accepted as a poster at the NeurIPS 2026 Workshop TAE
(Trust-AI-Eval). Code and data: \url{https://github.com/Ankush-kadu/activation-patching-reachable-state-geometry}.}\endgroup

% ============================================================== 1 ====
\section{Introduction}
\label{sec:intro}

Activation patching is a central experimental tool of mechanistic interpretability, used to locate factual associations
\citep{meng2022locating}, analyse bias mediation \citep{vig2020gender}, trace circuits
\citep{wang2022ioi,hanna2023greater} and automate circuit discovery \citep{conmy2023acdc}. An experimenter runs the model on a prompt, replaces
the output of one component, a \emph{site}, with its output on another prompt, and records how a readout changes
\citep{zhang2024patching,heimersheim2024patching}. In circuit analyses the readout is often a logit difference, the
logit of the correct answer minus that of a competing one \citep{wang2022ioi}. Suppose that replacing the output of
head $X$ moves this logit difference by $+0.5$. A circuit analysis would like to reuse that number when it describes
the rest of the mechanism, which assumes that the effect carries over. Would head $X$ still contribute $+0.5$ if an
upstream head changed? In general, single-site patching cannot answer this question. This paper explains why, characterises which
interactions it can miss, and shows how to measure the change directly.

The question matters because circuit explanations are assembled from such measurements and then reused. The
indirect-object-identification (IOI) circuit in GPT-2 small \citep{wang2022ioi} was traced with path patching, which patches along specific paths between heads, and
related analyses of how S-Inhibition heads influence Name Mover heads. The published circuit later served as a
reference for evaluating automated circuit discovery \citep{conmy2023acdc}, and the IOI task is part of an
interpretability benchmark \citep{mueller2025mib}. Whenever a single-head measurement is used to predict the effect of
a change elsewhere in a circuit, it matters whether the experiment could have detected an interaction between the two
heads. If it could not, a mechanism can fit every single-site measurement and still mispredict what happens when two
heads change together. Recent work calls for interpretability claims to state what their interventions can identify
\citep{joshi2026causality,lin2026disclose}. This paper provides that analysis for activation patching, together with a
direct test of the assumption.

\paragraph{Contributions.} (i) A framework that determines, from the recorded intervention states alone and before
any readout is scored, which candidate mechanisms an experiment can distinguish (\S\ref{sec:setup}). (ii) Exact
results on idealised two-site slices: the operator on the connection decides which polynomial interactions single-site
patching cannot detect, for affine, normalisation, softmax and gated connections. On affine connections, single-site
patching cannot detect $\KBlindTwo\%$ of the dimensions of the cubic hypothesis class with two sites and
$\KBlindFive\%$ with five (\S\ref{sec:theory}). (iii) A measurement across $\MatchedN$ screened captures, each one run
for a pair of sites and a seed, in GPT-2 small, Qwen3-8B and Qwen3.5-9B. Single-site patching is never exactly blind in
the tested cubic class. Along the interaction predicted by the connection, however, a hypothesis difference is more
than ten times larger in mean square on held-out joint states than on single-site states in $\PredMTotalCount$
captures, and this ratio exceeds the 95th percentile of the same capture's $200$ random-direction ratios in
$\HeldoutBeatsNull$ (\S\ref{sec:practical}). (iv) A direct, chart-free test of whether a head's effect carries over,
obtained by executing the two interventions that single-site patching omits. On the S-Inhibition $\to$ Name Mover edge
of the IOI circuit, with the Name Mover's output fixed, the effect changed by a median of only
$\AdqRmsTPctEffect\%$ of its typical size (root mean square over each run's tested states), and a preregistered
replication on $\RepPrompts$ new prompts met both size criteria in all $\RepTemplates$ templates (\S\ref{sec:ioi}).
(v) Two pitfalls in comparing patching designs, each with a remedy, and a checklist for reporting patching evidence
(Sections~\ref{sec:pitfalls} and~\ref{sec:limitations}). The trained-model analysis is a post hoc re-analysis and the
comparison with control heads is unresolved, so we treat the transport result as exploratory.

We illustrate the central idea with two heads. Once the candidate mechanisms are fixed, the states an experiment
reaches determine which pairs of them it can distinguish, before any readout is seen. The readouts then show how well each hypothesis fits
the observations.

\begin{figure}[t]
\centering
\includegraphics[width=\textwidth]{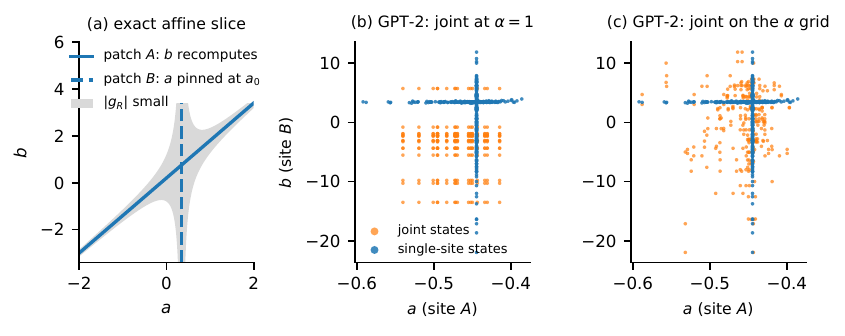}
\caption{Where single-site patching goes. \emph{(a)} On an exact affine edge, single-site states lie on two arms
(blue), where the interaction $g_R$ is zero; the grey region is where $|g_R|$ is small. \emph{(b)} One GPT-2 small
capture: attention outputs at layers $5$ and $10$, seed $0$, unembedding chart before whitening. This is a block-level
pair, not the head-level IOI edge of \S\ref{sec:ioi}. Single-site states (blue) form a pinned arm and a thin, almost
horizontal recompute arm, because the edge is nearly flat in this chart. Joint states at interpolation coefficient (the fraction of the way from clean to donor output, \S\ref{sec:setup})
$\alpha=1$ (orange) lie off both. \emph{(c)} Putting the joint family on the single-site coefficient grid widens its
range (\S\ref{sec:pitfalls}).}
\label{fig:reachable}
\end{figure}

Take an \emph{edge} $A\to B$, a pair of sites where $A$'s output feeds $B$, with $B$ in the role of head $X$. Summarise each site's output by one number, $a$ and $b$; we call this pair of
coordinates a \emph{chart}, a coordinate system for the two sites. A mechanism hypothesis predicts the readout at each point of the chart. Patching $A$ lets $B$ recompute;
on an affine edge the states move along a line $b=ma+c$, the \emph{recompute arm}. Patching $B$ leaves $A$ at its
clean value $a_0$, which gives the \emph{pinned arm}; we write $b_0$ for the clean value of $B$. Single-site patching
therefore visits only these two lines, or \emph{arms}
(Figure~\ref{fig:reachable}a). Take any candidate mechanism and add a term that is zero at every state these
arms visit. The new mechanism makes the same predictions on all single-site data, so single-site data cannot tell the
two apart. One such term is constructed as follows. Add $\lambda g_R$, with $g_R(a,b)=(a-a_0)(b-ma-c)$, to any hypothesis $f$. The
first factor is zero on the pinned arm and the second is zero on the recompute arm. The predictions do not change
anywhere single-site patching goes, so no amount of single-site data can detect this term. Yet $f$ and $f+\lambda g_R$
disagree about the opening question. Relative to $f$, the effect of moving $B$ from $b_0$ to $b$ changes by
$\lambda(a-a_0)(b-b_0)$ when $A$ moves from $a_0$ to $a$.

The term $g_R$ is an interaction: it is nonzero only when $A$ moves and $B$ sits off the line it would recompute to.
Patching both sites at once, \emph{joint} patching, reaches states off the arms where $g_R\neq0$. \emph{Clamping},
which patches $A$ while holding $B$ at its clean output, also reaches such states unless the edge is flat ($m=0$). To test whether $B$'s effect carries over once $A$ has
changed, change $A$ and take two readouts: one with $B$ forced to a replacement output and one with $B$ held clean.
Single-site patching generally supplies neither. Section~\ref{sec:ioi} runs both on a published circuit. In a
trained model the arms are not exact lines. Figure~\ref{fig:reachable}b shows one GPT-2 small capture. The pinned arm
is a line, but the recompute arm thickens into a thin band, so $g_R$ is small rather than zero on single-site states.

\paragraph{Related work.} The underlying mathematics is classical. Polynomial hypotheses that agree at every design point are
\emph{aliased} \citep{pistone1996groebner}, designs can be compared by the support preorder of their information
matrices \citep{pukelsheim2006optimal}, and recompute versus clamp parallels total versus controlled direct effects
\citep{pearl2001direct}. What is new under patching is who chooses the design points. The analyst sets the replacement
at $A$, but the network decides where $B$ lands, so the design must be read off the recorded states. Guides to patching
cover metrics, corruptions and path patching
\citep{zhang2024patching,heimersheim2024patching,goldowskydill2023pathpatching}; we ask instead what a whole family of
interventions can exclude. Mechanistic explanations are not unique \citep{meloux2025everything}, and subspace patches can
act through dormant pathways \citep{makelov2024illusion}. Our transport contrast
parallels the interaction terms of mediation analysis \citep{vanderweele2014fourway}. Appendix~\ref{app:related}
discusses further work.

% ============================================================== 2 ====
\section{What a patching experiment can tell apart}
\label{sec:setup}

The example generalises. Fix a model and $k$ internal sites. A chart summarises their outputs as a state
$z\in\mathcal Z$, and a scalar readout $y$ records behaviour; throughout, $y$ is a logit difference. Each intervention
$q$ in an \emph{intervention family} (or \emph{design}) $Q$ produces a state $z(q)$, and together these form the family's \emph{reachable
set} $R_Q=\{z(q):q\in Q\}$.  A \emph{mechanism hypothesis} is a function $f:\mathcal Z\to\mathbb R$ that predicts the
readout. A hypothesis here is coarser than an algorithm: two algorithms that make the same predictions in the chart count as one. Functions that agree on finitely many states can differ arbitrarily elsewhere, so we fix a \emph{mechanism
class} $\mathcal F$: a linear space of the hypotheses the analyst is willing to consider. Two hypotheses can then be
told apart by $Q$ exactly when they disagree somewhere on $R_Q$, that is, when their difference is not zero there.

\begin{definition}[Blind space; resolving preorder]
\label{def:blind}
The evidence map $\mathrm{Res}_Q:\mathcal F\to\mathbb R^{R_Q}$, $f\mapsto f|_{R_Q}$, restricts each hypothesis to the
reachable set. Define
\[
  N(Q,\mathcal F)=\ker\bigl(\mathrm{Res}_Q|_{\mathcal F}\bigr),\qquad
  Q_1\succeq_{\mathcal F}Q_2 :\iff N(Q_1,\mathcal F)\subseteq N(Q_2,\mathcal F).
\]
Two hypotheses are indistinguishable under $Q$ exactly when their difference lies in the \emph{blind space}
$N(Q,\mathcal F)$. So $Q_1\succeq_{\mathcal F}Q_2$ means that $Q_1$ distinguishes every pair that $Q_2$ distinguishes.
Families are \emph{incomparable} when each distinguishes a pair the other cannot.
\end{definition}

On the affine edge of \S\ref{sec:intro}, single-site patching cannot tell $f$ from $f+\lambda g_R$: their difference
$\lambda g_R$ vanishes on both arms, so it lies in the blind space.

The definition uses only the geometry of the reachable set and the class, never the readouts, so designs can be compared before an
experiment is scored. Pooling evidence can only shrink the blind space, since
$N(Q_1\cup Q_2,\mathcal F)=N(Q_1,\mathcal F)\cap N(Q_2,\mathcal F)$.

Our interventions replace a site's whole output. A donor prompt and a coefficient $\alpha$ set the output a fraction
$\alpha$ of the way from its clean value ($\alpha=0$) to its value on the donor ($\alpha=1$). Values outside $[0,1]$ extrapolate. On a two-site edge, write $a_0,b_0$ for the clean coordinates and $b=e(a)$ for where $B$ lands when $A$
is patched. The three families of the example then have simple reachable sets. Single-site patching $S$ reaches
$\{(a,e(a))\}\cup\{(a_0,b)\}$, clamping reaches $\{(a,b_0)\}\cup\{(a_0,b)\}$, and joint patching sets both outputs at
once. The relation $b=e(a)$ is an idealisation; for trained models we use the recorded states directly
(\S\ref{sec:practical}).

% ============================================================== 3 ====
\section{On idealised edges, the operator decides when blindness starts}
\label{sec:theory}

This section shows that, on idealised two-site slices, whether single-site patching can miss an interaction depends
on the function that carries one site's output to the next. Through an affine connection, single-site patching can miss even a degree-$\DstarAffine$ product term; through a
normalisation it can miss only terms of degree $\DstarRMSNorm$ or more; and through a softmax or gate it misses no
polynomial term. We now make this
precise. We call the map $e$, which sends $A$'s
coordinate to where $B$ lands, the edge's \emph{operator}; it may be affine,
a normalisation or a softmax. Which interactions can hide becomes a question about polynomials: a polynomial term hides exactly when it is zero at
every state single-site patching reaches. Take the mechanism class to be the polynomials in $(a,b)$ of total degree at most $d$, written $\mathcal P_{\le d}$.
The blind space is then the set of polynomials of degree at most $d$ that are zero at every state the experiment
reaches. Formally $N(Q,\mathcal P_{\le d})=I(R_Q)\cap\mathcal P_{\le d}$, where the \emph{vanishing ideal} $I(R_Q)$
is the set of all polynomials that are zero on $R_Q$. We call the smallest degree with a nonzero blind
polynomial the \emph{onset}, $d^*(Q)=\min\{d:N(Q,\mathcal P_{\le d})\neq0\}$. Proofs are in Appendix~\ref{app:proofs}.

On the affine edge $e(a)=ma+c$, with $b_0=ma_0+c$, a polynomial that vanishes on both arms must contain the factor
$a-a_0$ and the factor $b-ma-c$. Every blind polynomial is therefore a multiple of the \emph{generator} $g_R$, and
blindness starts at degree $\DstarAffine$. Clamping swaps the recompute arm for the line $b=b_0$, so its generator is
$g_C=(a-a_0)(b-b_0)$; on this edge $g_C-g_R=m(a-a_0)^2$. The two designs impose the same number of constraints at every
degree. Yet when $m\neq0$ and $d\ge2$, each sees a difference the other misses
(Proposition~\ref{prop:incomparable} in Appendix~\ref{app:proofs}), so counting constraints does not rank designs.

The same argument works whenever the recompute arm lies on a curve $P=0$. The product $(a-a_0)P$ vanishes on both arms,
so blindness starts no later than degree $1+\deg P$. Theorem~\ref{thm:operator} gives the conditions under which it
starts exactly there.

\begin{theorem}[Operator-slice theorem]
\label{thm:operator}
On a two-site slice, let the pinned branch be $A_0=\{(a_0,b):b\in B_0\}$ with $B_0$ infinite. Let the recompute branch
$\Gamma=\{(a,e(a))\}$ be an infinite subset of the zero set $V(P)$. Assume that $P\in\mathbb R[a,b]$ is irreducible
(not a product of two nonconstant polynomials) and not a constant multiple of $a-a_0$. Then
$I(A_0\cup\Gamma)=\bigl((a-a_0)P\bigr)$ and $d^*=1+\deg P$.
\end{theorem}

Affine edges have $\deg P=\DstarAffineDegP$, so $d^*=\DstarAffine$. For the normalisation slices
$e(a)=\beta+\omega a/\sqrt{\kappa a^2+\epsilon}$ with $\omega\neq0$, $\kappa>0$ and $\epsilon>0$, clearing the radical
gives an irreducible $P$ of degree $\DstarRMSNormDegP$. The pinned arm's factor $a-a_0$ adds one more degree, so blindness starts at
$d^*=\DstarRMSNorm$ on both the analysed RMSNorm and LayerNorm slices. For two-key softmax and
sigmoid-gated linear-attention slices with nonzero parameters, no nonzero polynomial vanishes on the recompute arm (a
separate lemma). Their blind spaces are zero at every finite degree, so $d^*=\DstarSoftmax$ for both.
Figure~\ref{fig:regimes} (left) checks the affine, RMSNorm and softmax cases numerically. Which rival hypotheses single-site patching cannot exclude therefore depends on the operator.
Section~\ref{sec:practical} shows that, for the recompute single-site and matched joint designs in trained models, the
practical issue is weak evidence rather than exact blindness.

Real circuits involve more than two heads, so we ask how
blindness grows with the number of sites. With $k$ sites patched one at a time, the reachable set is $k$ curves
through the clean state. On arm $i$ the patched coordinate is swept, upstream coordinates stay clean and downstream ones
recompute. When every coordinate on arm $i$ is a polynomial in the swept coordinate, write $\eta_i$ for the highest degree among
them; $\eta_i=1$ for a straight arm. A degree-$d$ hypothesis then becomes a one-variable
polynomial of degree at most $d\eta_i$ on that arm.

\begin{proposition}[Restriction-rank bound for polynomial single-site arms]
\label{prop:ksite}
Let the coordinates on arm $i$ be polynomials $z_{ij}(t)$ of the swept coordinate $t$, with $z_{ii}(t)=t$, and let all
arms pass through the clean configuration. Set $\eta_i=\max\bigl(\{1\}\cup\{\deg z_{ij}:z_{ij}\text{
nonconstant}\}\bigr)$. Then $\operatorname{rank}(\mathrm{Res}_Q|_{\mathcal P_{\le d}})\le1+d\sum_i\eta_i$, so
$\dim N(Q,\mathcal P_{\le d})\ge\binom{d+k}{k}-\bigl(1+d\sum_i\eta_i\bigr)$.
\end{proposition}

The degrees must include all downstream recomputation (Appendix~\ref{app:proofs}). For affine edges $\eta_i=1$ and the
bound $1+kd$ is attained, while the class has $\binom{d+k}{k}$ dimensions. At degree $3$ the blind space takes up
$\KBlindTwo\%$ of the class's dimensions with two sites and $\KBlindFive\%$ with five. On affine edges, single-site evidence therefore covers a shrinking fraction of the hypothesis class as the number of
sites grows, whereas a joint product grid with at least four distinct values on each axis resolves the whole degree-$3$
class.

\begin{remark}[Scope of the exact results]
\label{rem:scope}
The onset results concern the continuum of states on idealised two-site slices, and
Proposition~\ref{prop:ksite} concerns $k$ polynomial arms; both are stated in the chosen polynomial coordinates. Every
finite grid admits some vanishing high-degree polynomial, so numerical checks can support them but never prove them.
They say nothing about whether a full transformer has exactly blind mechanisms.
\end{remark}

% ============================================================== 4 ====
\section{In the tested runs, the edge predicts an interaction that single-site patching constrains weakly}
\label{sec:practical}

This section moves from idealised edges to recorded activations. In all $\MatchedN$ screened runs, single-site
patching could distinguish every pair of hypotheses in the cubic class we tested. Even so, the recorded states let us
name, from the edge alone, an interaction that joint patching usually constrained much more strongly.

\paragraph{Setup.} We study $\MatchedN$ two-site captures from GPT-2 small, Qwen3-8B and Qwen3.5-9B on five tasks,
forming $\PredCells$ model--task cells (Appendix~\ref{app:protocol}). A capture is one run for a pair of attention
outputs and a seed. Each site is summarised by one coordinate, its projection on the unembedding difference of the
task's two answer tokens, rescaled to unit variance (whitened). The mechanism class is the cubic polynomials in these
two coordinates. In trained models the recompute states scatter around any fitted curve, and in every capture both the
recompute single-site design $S$ and the matched joint design $J$ have full numerical rank on the recorded states in
this class, so neither is exactly blind to any direction.\footnote{Clamping differs. Each of its arms pins one
coordinate, which gives it an exact polynomial kernel by construction (Appendix~\ref{app:protocol}).} Under the exact
preorder of \S\ref{sec:setup} the two designs therefore tie, and the useful question becomes which directions each
design constrains only weakly.

\paragraph{Measuring weak evidence.} A difference between two mechanisms is hard to detect if it stays small wherever
the experiment goes. We therefore measure the \emph{energy} of a hypothesis difference $h$: its mean square over the
states a family reaches. In matrix form, with a feature map $\varphi$ spanning the class and a weighting $\mu_Q$ of the
family's states, $h=\theta^\top\varphi$ has energy $E_Q(h)=\mathbb E_{\mu_Q}[h^2]=\theta^\top G_Q\theta$, where
$G_Q=\mathbb E_{\mu_Q}[\varphi\varphi^\top]$ is the Gram matrix. Zero energy is exact blindness: when $\mu_Q$ is
supported on the whole reachable set, the vectors in $\ker G_Q$ represent exactly the blind space, since our polynomial
features are continuous with finite second moments. We compare the two families by the \emph{held-out $J/S$ energy
ratio} $r(h)=E_J(h)/E_S(h)$, fixing each direction on one half of each design's rows and scoring it on the other half.
At $r=10$, the mean squared gap between two mechanisms is ten times larger on joint than on single-site states. Because
$r$ is unchanged when $h$ is rescaled, a large $r$ does not imply a large gap in logits. The joint family uses the same
coefficient grid as the single-site family, for reasons given in \S\ref{sec:pitfalls}. We exclude captures whose joint
readouts barely vary ($\mathrm{sd}<0.15$), since they carry little signal. This \emph{readout screen} is the only use of
readouts in the geometry analyses of Sections~\ref{sec:practical} and~\ref{sec:pitfalls}. The counts below describe
screened captures, which share prompts and tasks, so we make no population claim.

\paragraph{The weak direction can be named from the edge.}
\label{sec:predict}
On an affine edge each arm supplies one factor of $g_R=(a-a_0)(b-ma-c)$, so $g_R$ can be estimated from intervention
states alone: $a_0$ from the arm where $B$ is patched, and $(m,c)$ from a least-squares fit of $b$ on $a$ along the arm
where $A$ is patched, both on training rows. No readout enters, and the direction is not chosen to separate the
families. Its held-out ratio exceeds ten, one order of magnitude, in $\PredMTotalCount$ captures. Because random
directions can also reach ratios near ten, we also compare $g_R$ with each capture's own random baseline: in
$\HeldoutBeatsNull$ captures its ratio exceeds the 95th percentile of $200$ random directions in that capture (cell
medians of that percentile range from $\HeldoutNullLo$ to $\HeldoutNullHi$).

\paragraph{Modelling the edge adds to the classical interaction.} The classical one-at-a-time interaction
$g_C=(a-a_0)(b-b_0)$ ignores the edge. On the same states it exceeds ten in $\OfatRowCount$ captures, against
$\PredMTotalCount$ for $g_R$, a net gain of $\OfatNetDifference$ captures from modelling the realised edge (counts at
one threshold; the overlap is in Appendix~\ref{app:matched}). The classical term performs well because many edges are
nearly flat in this chart: recomputing $B$ barely moves it from its clean value. We measure edge strength as the ratio
of the spread of $b$ when $A$ is patched to its spread when $B$ is patched. On stronger edges $r(g_R)$ falls (Spearman
$\EdgeBlindRho$ over $\EdgeN$ captures), but its advantage over $g_C$ grows, so modelling the edge matters most where
the edge is strong (Appendix~\ref{app:matched}).

\paragraph{Robustness.} The captures were collected and inspected before this analysis was fixed, so it is a post hoc
re-analysis. Two stricter checks agree (Appendix~\ref{app:units}). Holding out whole donor indices rather than rows,
$g_R$ still exceeds ten in $\DonorNamedCount$ captures, although that split does not fully separate donor content (in
$\AuditDonorDupCaptures$ captures a held-out donor's patch-$A$ trajectory duplicates a training donor's). Grouping
captures by cell and prompt, every capture clears ten in $\HeldoutPromptStrict$ groups. Appendix~\ref{app:robust}
checks the estimator on synthetic data, and Appendix~\ref{app:three} adds a third site.

\paragraph{Implications.} The single-site arms alone were enough to name $g_R$, yet single-site states pin it down far
less than joint states do. This matters for head $X$: $g_R$ is exactly the kind of term that changes a head's effect
after an upstream change (\S\ref{sec:intro}). These results concern how visible a candidate interaction would be, for
pairs of attention-layer outputs in one chart; they do not say whether the network departs from additivity.
Section~\ref{sec:ioi} measures that departure directly, on specific heads and without a chart. That measurement cannot
tell $g_R$ from $g_C$, since on the affine slice the two change the measured contrast by the same amount.

% ============================================================== 5 ====

\section{Does the Name Mover's effect carry over? Executing the missing corners}
\label{sec:ioi}

We now return to head $X$. Does its measured effect carry over when an upstream head changes? If it does, we say
that the effect \emph{transports}. We test this on the S-Inhibition $\to$ Name Mover edge of the IOI circuit
of \citet{wang2022ioi} in GPT-2 small, with the Name Mover in the role of head $X$. The circuit completes prompts such as ``When Mary and John went to
the store, John gave a drink to'' with ``Mary''. In it, Name Movers copy the remaining name to the output, and
S-Inhibition heads make them attend less to the repeated name. We force the Name Mover's output,
so the test asks only how the readout combines the two heads' outputs on the executed states. It does not test the
query pathway by which S-Inhibition changes what the Name Mover writes, and it does not contradict the mechanism that
\citet{wang2022ioi} found by path patching.

\begin{figure}[t]
\begin{tcolorbox}[title={The transport contrast}]
Let $A$ be the S-Inhibition head and $B$ the Name Mover. Write $u$ and $v$ for replacement outputs of $A$ and $B$, and
$0$ for each site's clean output. $Y(u,v)$ is the readout with $A$'s output set to $u$ and $B$'s output forced to $v$,
so $Y(u,0)$ clamps $B$ at its clean output. Patching $B$ alone measures its effect \emph{in the context} where $A$ is
clean,
\[
  \Delta_B(v)=Y(0,v)-Y(0,0).
\]
If this effect transports, the same $\Delta_B(v)$ should apply once $A$ is changed to $u$. Exactly, and without any
model,
\[
  Y(u,v)-Y(u,0)=\Delta_B(v)+T_C(u,v),\qquad T_C(u,v)=Y(u,v)-Y(u,0)-Y(0,v)+Y(0,0).
\]
The four-point \emph{transport contrast} $T_C$ is zero at every tested $(u,v)$ exactly when $B$'s measured effect
carries over to every tested state of $A$. Patching $A$ alone observes $Y(u,v_{\mathrm{rec}}(u))$, where
$v_{\mathrm{rec}}(u)$ is the output $B$ recomputes to; it generally observes neither the joint corner $Y(u,v)$ nor the
clamped corner $Y(u,0)$. Single-site evidence therefore does not determine $T_C$ without further assumptions.
Executing the two missing corners measures it directly.
\end{tcolorbox}
\end{figure}

The transport contrast $T_C$ defined in the box is zero for any readout that adds separate effects of the two sites, so it measures departure from
additivity at the executed corners. It also links back to the blind directions. On the affine slice of
\S\ref{sec:theory}, adding the single-site-blind term $\lambda g_R$ to a hypothesis changes its contrast by
$\lambda(a-a_0)(b-b_0)$, and adding $\lambda g_C$ changes it by the same amount. Two mechanisms that agree on every
single-site state can therefore disagree about $T_C$, and executing the corners decides between them without any
chart. The measured $T_C$ does not say which functional form the network uses.

\paragraph{Design.} We pair four S-Inhibition heads with three Name Movers, giving twelve published head pairs.
Controls replace the S-Inhibition heads with four early heads that have no claimed IOI role, paired with the same Name
Movers. A head pair on one prompt is a \emph{capture}. On three base prompts, $\QCN$ of the $36$ published captures and
$\QCCtrlN$ of the $36$ control captures passed the readout screen of \S\ref{sec:practical} and ran without error (Appendix~\ref{app:protocol}). Each capture executes the single-site arms, the clamped arm
and $256$ joint states; Appendix~\ref{app:ioi} lists the heads.

\paragraph{On average, the Name Mover's effect carries over closely.} For each published capture we take the root mean
square (RMS) of $T_C$ over its joint states and compare it with two scales of the Name Mover's own effect
(Table~\ref{tab:transport}). The first is its \emph{B-arm swing}, the range of its effect over the tested
coefficients.\footnote{The swing is the readout range over the single-site $B$ arm, $32$ donors $\times$ $\NAlpha$
coefficients. Its median is $\MfSwingMed$ logits and it is never below $\MfSwingMin$ logits.} The swing includes
extrapolated coefficients, so it is larger than the second scale, the Name Mover's \emph{typical effect}: its mean
absolute per-donor effect at $\alpha\approx1$ (Appendix~\ref{app:ioi}). In the exploratory captures the median ratio
of swing to typical effect is $\AdqSwingOverEffect$. The median RMS is $\AdqRmsT$ logits. Dividing by each
capture's own swing or typical effect gives median percentages of $\AdqRmsTPct\%$ and $\AdqRmsTPctEffect\%$. The RMS averages over the joint states of each capture, so
single states can be larger. The median of the within-capture maxima of $|T_C|$ is $\AdqMaxTPctEffect\%$ of the typical
effect, and no exploratory contrast exceeds about $\MfMaxAbsMax$ logits. We chose this analysis after seeing the data,
so it is exploratory. We then registered a test of its two RMS criteria on fresh prompts.

\begin{table}[t]
\centering
\small
\caption{Transport contrasts with the Name Mover's output forced. Each capture's root
mean square (RMS) is taken over its $256$ joint states. Exploratory rows give medians over $\QCN$ screened published
captures on three base prompts, with percentages computed within capture. In the replication on $\RepPrompts$ fresh
prompts, the first two rows count published-edge template summaries, each the median over screened captures within a
prompt and then over prompts within the template. The next two rows count screened captures whose RMS exceeded ten
times their own re-execution floor (the float32 floor), the numerical noise measured by re-running the rows in reverse order with a different batch size.}
\label{tab:transport}
\begin{tabular}{llr}
\toprule
Run & Quantity & Value\\
\midrule
Exploratory & median RMS of $T_C$ & $\AdqRmsT$ logits\\
 & \quad as a share of the B-arm swing & $\AdqRmsTPct\%$\\
 & \quad as a share of the typical effect & $\AdqRmsTPctEffect\%$\\
 & median within-capture maximum $|T_C|$, share of typical effect & $\AdqMaxTPctEffect\%$\\
\midrule
Preregistered & templates with median RMS below $\MfMargin$ logits & $\RepHAbsCount$\\
replication & templates with relative RMS below $1\%$ & $\RepHRelCount$\\
 & captures above ten times the float32 floor, published & $\RepPubResolved$\\
 & captures above ten times the float32 floor, control & $\RepCtlResolved$\\
 & comparison with control heads & unresolved\\
\bottomrule
\end{tabular}
\end{table}

\paragraph{A preregistered replication agrees on size.} Before collecting new data, we registered a replication
(Appendix~\ref{app:replication}). It used $\RepPrompts$ fresh prompts, three from each of $\RepTemplates$ IOI templates
of \citet{wang2022ioi} that we had not used, with the same heads, controls, design and analysis. For each prompt we
take the median RMS over its screened published captures. Each template is then summarised by the median over its
three prompts. The absolute RMS was below $\MfMargin$ logits in $\RepHAbsCount$ templates. The RMS relative to the
B-arm swing was below $1\%$ in $\RepHRelCount$, with template medians from $\RepVrelMin\%$ to $\RepVrelMax\%$. Both
registered smallness criteria passed. We set both margins after the exploratory analysis. The absolute margin is more than ten times
the exploratory median RMS of $\AdqRmsT$ logits, so the relative margin is the more informative one. Because the swing was a median $\AdqSwingOverEffect$ times the
typical effect in the exploratory captures, the $1\%$-of-swing margin is looser in typical-effect units.
Individual screened captures can be larger: over screened published captures, the 90th percentile of the relative RMS
was $\RepPubRelPninety\%$ and the maximum $\RepPubRelMax\%$.

\paragraph{Comparison with control heads.} The replication measured the
float32 floor, the numerical noise of the computation, by re-executing every row in reverse order, with batch size $7$ instead of $16$ and the clean pass and
donor activations held fixed. Contrasts exceeded ten times their floor in $\RepPubResolved$ screened published captures
but in only $\RepCtlResolved$ screened control captures. Descriptively, the median control RMS was $\RepCtlRmsMed$ logits against $\RepPubRmsMed$ for published captures.
Most control contrasts were too small to clear ten times their float32 floor, so the registered comparison with
controls was floor-limited,
and we make no claim that the edge's contrast exceeds theirs. Even a resolved comparison would mix claimed role with
depth, since the control heads sit in layers 0--3 and the published S-Inhibition heads in layers 7--8. We report the published/control ratio
and the template log differences only descriptively (Appendix~\ref{app:replication}); they have no calibrated
interpretation. Because the preregistered rule required every
criterion to pass, the unresolved control comparison leaves the transport result exploratory as a whole.

\paragraph{A cubic model fitted on single-site data does not recover the missing corners.} Could a model fitted on single-site data
have predicted these contrasts? We tried a natural one: cubic polynomials in a fixed chart derived from the weights,
with interaction coefficients $\gamma$ on $ab$, $a^2b$ and $ab^2$. The chart follows the published claim that
S-Inhibition heads act on the Name Movers' \emph{query}, and neither of its directions uses intervention states or
readouts. Single-site data inform $\gamma$ only through the scatter of $b$ around the
fitted edge, and the fit fails on the exploratory captures. On the first donor partition, the fit's median held-out RMS error in
predicting $T_C$ is $\MfSThree$ logits, against $\MfZero$ for simply predicting zero. On joint readouts the full cubic generalises very poorly, with held-out $R^2$ of $\RespSJR$, whereas an
additive-only refit reaches $\RespAddThreeSJ$. For each of the two $R^2$ values, we take the median over captures
within each of ten donor partitions, then the median over partitions. Using its nominal $95\%$ confidence region, the full single-site cubic gave an upper limit for the size of the
contrast at each state. The executed $|T_C|$ stayed within those limits at all $256$ joint states in only
$\AdqCoverUpperS$ screened published captures of the exploratory run
(Figure~\ref{fig:scales} and Table~\ref{tab:scales} in Appendix~\ref{app:ioi}). Under a pooled covariance the fit rejects
additivity ($\gamma=0$) in $\QCRejS$ published and $\QCCtrlRejS$ control captures, which would appear to be a detection. Under five alternative covariances, however, the controls reject in $\DetCtrlRobLo$ to $\DetCtrlRobHi$ of
$\QCCtrlN$. A single-site rejection of $\gamma=0$ concerns the arms only; it does not measure $T_C$ at the joint corners.

\paragraph{Implications for head $X$.} Our scope is GPT-2 small, the published S-Inhibition $\to$ Name Mover head pairs,
output replacement with the Name Mover's output forced, and the tested grid of states. Within it, reusing the Name
Mover's single-site effect after S-Inhibition changes is off by a median RMS of $\AdqRmsTPctEffect\%$ of its typical
effect over the exploratory captures, and single states can be off by more. On these states, in a typical exploratory
capture, the readout combines the two heads' forced outputs close to additively. The effect of S-Inhibition with the Name Mover clamped, plus
the Name Mover's own effect, nearly gives their joint effect. This is not the same as adding two ordinary single-site
patches, since patching S-Inhibition alone lets the Name Mover recompute.
Whether S-Inhibition changes what the Name Mover writes is a separate question that this design does not test.
Measuring the change took two interventions per state that single-site patching does not run.

% ============================================================== 6 ====
\begin{figure}[t]
\centering
\includegraphics[width=\textwidth]{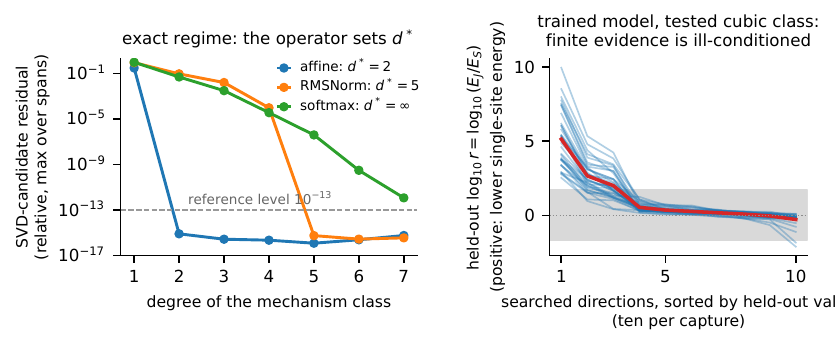}
\caption{Exact slices and a trained model. \emph{Left (\S\ref{sec:theory}):} on each operator slice, a singular value
decomposition fits a candidate polynomial relation on $[-2,2]$, $[-8,8]$ and $[-32,32]$; each curve shows the largest
relative residual across the three spans. The affine and RMSNorm residuals reach numerical precision at their proved
onsets. The softmax residual keeps falling, but residuals alone cannot tell an exact relation from a close
approximation (Remark~\ref{rem:scope}). The dashed line marks $10^{-13}$. \emph{Right (\S\ref{sec:pitfalls}):} GPT-2
small IOI captures on the matched coefficient grid. Each thin line orders one capture's searched directions by
held-out $\log_{10}r=\log_{10}(E_J/E_S)$, from high to low; positive values mean lower single-site energy. The thick line
averages across captures at each rank. The grey band is the same-family envelope, the spread of scores between two
halves of the single-site family, shown for display; escape counts use a leave-one-capture-out band
(Appendix~\ref{app:protocol}).}
\label{fig:regimes}
\end{figure}

\section{Two pitfalls in comparing intervention families, and their remedies}
\label{sec:pitfalls}

Section~\ref{sec:practical} scored one direction named from the edge. Comparing two designs over the whole class
requires a search, in each capture, for the directions each design constrains least. This section identifies two ways
such a search can mislead, each with a simple remedy. Differences in intervention range can distort a comparison, as can choosing a chart using
the same states later used to score it. First, a joint family built from raw donors can look weaker than
single-site patching because the single-site grid extrapolates; matching the coefficient grids removes most of this
effect. Second, a chart fitted on the states it later scores can make two designs look complementary; the remedy is to
fit the chart on separate data.

\paragraph{The search.} For each capture we look for the hypothesis differences that one design sees much less than
the other. These are the solutions of a generalised eigenvalue problem on the two Gram matrices, solved on training
rows:
\[
  \bigl(G_S+\lambda G_{\mathrm{ref}}\bigr)w=\nu\,\bigl(G_J+\lambda G_{\mathrm{ref}}\bigr)w,\qquad
  G_{\mathrm{ref}}=\tfrac12(G_S+G_J),\quad\lambda=10^{-6}.
\]
Its eigenvalues and eigenspaces do not depend on the feature basis (Appendices~\ref{app:protocol}
and~\ref{app:noninvariant}). We score each eigenvector on held-out rows by its ratio $r=E_J/E_S$, as in
\S\ref{sec:practical}, and sort the scores into the capture's \emph{profile} (Figure~\ref{fig:regimes}, right). The
high end of the profile holds directions that single-site evidence constrains least relative to joint evidence, the role
$g_R$ played in \S\ref{sec:practical}; the low end holds the reverse.

Even when both halves come from the same family, chance differences between them can produce extreme ratios. As a
reference we therefore run the same procedure on two halves of $S$; the spread of their profiles is the \emph{same-family envelope}. An end of a
profile \emph{escapes} when it lies beyond this envelope. An escape at the low end, below the leave-one-capture-out
lower bound, is a \emph{reverse escape}: a direction that joint evidence constrains less than single-site evidence
does. The halves are smaller than the full families and captures share prompts, so escape counts are descriptive and
any $p$-values nominal (Appendix~\ref{app:protocol}).

\paragraph{Pitfall 1: unmatched intervention ranges.} When joint states use raw donors, $\EdgeFrozenEscapes$ captures show a reverse escape, apparently a direction that
joint patching constrains less than single-site patching does. The number of reverse escapes depends strongly on the coefficient grid. Our
single-site family sweeps $\alpha$ from $-0.5$ to $1.5$, half a clean-to-donor distance beyond each end, while raw donors
($\alpha=1$) never extrapolate. The wider grid can take single-site states into regions where some polynomials have much larger values, raising
their mean squared value under single-site patching for reasons unrelated to any mechanism. The same-family envelope
cannot catch this, because both halves of $S$ share one range. Recollecting the joint family on the single-site
coefficient grid, with the same prompts, donors and weights, reduces the count to $\EdgeMatchedEscapes$. Removing
single-site extrapolation alone, with the raw-donor joint states held fixed, leaves $\EdgeInteriorEscapes$, and
restricting both families to $\alpha\in[0,1]$ leaves $\EdgeSymmetricEscapes$; the few that remain go with stronger edges
(Appendix~\ref{app:matched}). The matched joint design is sparse rather than a full product grid, so even these
supports are not matched exactly.

\paragraph{Pitfall 2: a chart chosen on the scored states.} The second pitfall concerns the two single-site designs,
recompute and clamped. Recompute measures $A$'s total effect, including its route through $B$, while clamping holds $B$
fixed and measures a controlled direct effect. The two differ only where patching $A$ moves $B$, and in the unembedding
chart the edge is nearly flat (median absolute whitened slope $\RCUchartChartM$), so they almost coincide. A natural
step is to choose a chart in which the edge is steep. For each capture we chose the downstream coordinate that
maximises a regularised ratio of the variance of $b$ on the $A$ arm to its variance on the $B$ arm, using states but no
readouts, and fitted it on the first half of each arm, including some rows later used for scoring (Appendix~\ref{app:chart}). After dropping the
$\RCDropped$ captures with $\mathrm{sd}(b\mid\text{arm }A)/\mathrm{sd}(b\mid\text{arm }B)>50$, $\RCPooledR$ profiles
escape the envelope at the end where recompute evidence is weaker, and every one does at the end where clamped evidence
is weaker. Taken at face value, each design would constrain a direction the other misses.

Two observations caution against this reading. First, with the chart fitted on another seed of the same layer pair,
the recompute-weaker escape rate was $\RCXseedPooledR$. That rate comes from a different cohort, so the two rates are
not paired, and the content is not fully disjoint ($\AuditXseedSamePrompt$ of the cross-seed captures share the chart
source's base prompt, and donors come from a common pool). Second, the chart rests on little data: for GPT-2, the
denominator of its variance ratio is estimated from at most $16$ donor directions in a $768$-dimensional space, and in
the GPT-2 captures where this was logged the chosen direction was almost orthogonal to their span, so the regulariser
rather than the data controlled the ratio. The escape rates are consistent with chart selection inflating the apparent
gap, even when no readout is used, and cannot say how large the real gap between recompute and clamped evidence is.

% ============================================================== 7 ====
\section{Recommendations for reporting patching evidence}
\label{sec:limitations}

The results above translate into concrete reporting practice for any patching study.

\begin{tcolorbox}[breakable,title={An audit checklist for patching evidence}]
\begin{enumerate}[leftmargin=1.4em,itemsep=1pt,topsep=1pt,parsep=0pt]
\item Declare the mechanism class and the chart before using readouts to tune either. Record when the class, chart,
direction and decision criteria were fixed relative to data inspection.
\item Record the intervention \emph{states} as well as the readouts. They fix which hypotheses an experiment can
separate before any readout is scored (\S\ref{sec:setup}). Inspect each family's Gram matrix and its low-energy
directions (Appendix~\ref{app:protocol}).
\item Match intervention ranges before comparing families, because an extrapolating grid gives some polynomials energy
for reasons unrelated to any mechanism (\S\ref{sec:pitfalls}). If exact matching is infeasible,
repeat the comparison on the non-extrapolating states as a sensitivity check and report the remaining differences in
coefficient and donor coverage. Fit any data-derived chart on prompts and donors distinct from those used for scoring.
\item Test a working model on held-out executed contrasts before trusting its intervals, and keep hypothesis rejections
separate from effect estimates.
\item For a transport claim, declare the tested states, the aggregation unit and the margins. Execute the four-point
contrast, because single-site readouts generally leave it undetermined without further assumptions (\S\ref{sec:ioi}).
Measure its numerical floor by re-execution under a different batch composition. Declare a family-level resolution
criterion, and report the comparison as unresolved if either family fails it.
\item Account for unequal arm variances and donor dependence, and use inference units suited to the design: heads,
prompts or templates.
\end{enumerate}
\end{tcolorbox}

\paragraph{Scope and limitations.} The blind space depends on the class and the chart by construction, so the framework gives no
context-free answer. Full rank within a chart and class also does not show that those coordinates capture the internal
algorithm. The geometry analyses of Sections~\ref{sec:practical} and~\ref{sec:pitfalls} and the fitted IOI
cubic use one coordinate per site, a projection of a high-dimensional output, and a chart can itself discard an
interaction; the executed transport contrast needs no chart. The exact theorems cover idealised two-site slices and polynomial single-site arms in stated coordinates
(Remark~\ref{rem:scope} and Proposition~\ref{prop:ksite}). The results of \S\ref{sec:practical} describe weak identification in
the evaluated charts and classes. They come from a post hoc re-analysis of $\MatchedN$ screened captures in
$\PredCells$ cells that share prompts and tasks, with a joint design that is sparse rather than a full product grid,
and the named direction has not yet been validated on fresh captures.
Two separate registered hypotheses failed their criteria, and none of the results above relies on them
(Appendix~\ref{app:negatives}). The first was that a circuit frozen under one intervention family loses faithfulness
under the other in both directions. The second was that charts built from lens-style projections have distinctive
weakly constrained subspaces beyond matched compression controls. Table~\ref{tab:status} in Appendix~\ref{app:status} lists each claim with its
status.

\paragraph{Conclusion.} Before any readout is scored, the states a patching experiment reaches determine which
candidate readout functions it can distinguish in the chosen chart, and those states can be recorded in advance. This
turns a general concern about patching into a property an analyst can check before the analysis is run. On the
idealised edges we analyse, the operator decides which interactions single-site patching cannot detect, and on affine
edges the undetectable share grows with the number of sites. Across the $\MatchedN$ screened runs, single-site patching
had full numerical rank in the tested cubic class, yet it often constrained the edge-predicted interaction far more
weakly than joint patching. The transport contrast addresses this directly by measuring whether a head's effect carries
over. On the IOI edge, with the Name Mover's output fixed, a typical run showed that effect carrying over closely, and a
preregistered replication met both size criteria. The comparison with heads that have no claimed IOI role remains
unresolved, because most control contrasts did not clear the float32 floor by the registered margin. Recording
intervention states and executing the four-point contrast require only additional forward passes, two extra readouts per tested state, and together they
let a circuit claim state precisely what its patching evidence establishes.

\FloatBarrier
\bibliographystyle{plainnat}
\bibliography{refs}

\appendix
\section{Claims and their status}
\label{app:status}

Table~\ref{tab:status} lists the main claims with the status of the evidence for each.

\begin{table}[h]
\centering\footnotesize
\caption{Main claims and their status. \emph{Proved}: a theorem on the analysed slices. \emph{Exact}: a model-free
consequence of which states were executed. \emph{Post hoc}: a re-analysis of captures inspected before the analysis was
fixed. \emph{Exploratory}: an analysis chosen after seeing the data. \emph{Replicated}: a registered criterion met in
the preregistered replication. \emph{Floor-limited}: the controls did not meet the registered resolution criterion.
\emph{Failed}: a planned hypothesis whose criterion was not met.}
\label{tab:status}
\begin{tabularx}{\textwidth}{@{}>{\raggedright\arraybackslash}X >{\raggedright\arraybackslash}p{0.17\textwidth} l@{}}
\toprule
claim & status & where \\
\midrule
On analysed slices, in the stated coordinates, the operator on the edge fixes the single-site onset $d^*$ ($\DstarAffine$ affine, $\DstarRMSNorm$ RMSNorm, $\DstarLayerNorm$ LayerNorm, $\DstarSoftmax$ softmax, $\DstarGatedLinear$ gated) & proved & \S\ref{sec:theory}, App.~\ref{app:proofs} \\
For affine single-site designs at fixed $d\ge2$, the unresolved fraction grows with the number of sites & proved & \S\ref{sec:theory}, App.~\ref{app:proofs} \\
\midrule
On the matched grid, the held-out $J/S$ energy ratio exceeds $10$ for the readout-free named direction $g_R$ in $\PredMTotalCount$ captures, and for the one-at-a-time interaction $g_C$ in $\OfatRowCount$ & post hoc & \S\ref{sec:practical} \\
Unmatched ranges inflate reverse escapes ($\EdgeFrozenEscapes$ unmatched, $\EdgeMatchedEscapes$ matched); results consistent with chart selection on the scored states inflating the apparent gap & post hoc & \S\ref{sec:pitfalls} \\
\midrule
Single-site evidence alone does not determine the transport contrast $T_C$ & exact & \S\ref{sec:ioi} \\
The RMS of the executed $T_C$ is small on the tested grids (median over published captures, three base prompts) & exploratory & \S\ref{sec:ioi} \\
On $\RepPrompts$ fresh prompts from $\RepTemplates$ templates, template medians met both smallness criteria (absolute in $\RepHAbsCount$, relative in $\RepHRelCount$ templates) & replicated & App.~\ref{app:replication} \\
The cubic working model fitted to single-site data predicts $T_C$ poorly, and its intervals are miscalibrated & exploratory & \S\ref{sec:ioi} \\
Whether the IOI transport contrast exceeds that of heads with no claimed role & floor-limited & \S\ref{sec:ioi}, App.~\ref{app:replication} \\
\midrule
Reciprocal cross-family degradation of a frozen circuit; representation-specific weak subspaces & failed & App.~\ref{app:negatives} \\
\bottomrule
\end{tabularx}
\end{table}

% ===================================================================== B ====
\FloatBarrier
\section{Proofs and exact numerics}
\label{app:proofs}

Write $H_Q(d)=\operatorname{rank}(\mathrm{Res}_Q|_{\mathcal P_{\le d}})$ for the \emph{Hilbert profile}, the number of independent
constraints the evidence imposes. On an affine slice, recompute and clamped patching can constrain equally many
alternatives while missing different ones.
\begin{proposition}[Equal Hilbert profile, different blind spaces]
\label{prop:incomparable}
On the affine two-site slice with $e(a)=ma+c$ and $b_0=ma_0+c$, let $N_R$ and $N_C$ be the degree-$d$ blind spaces
of the recompute and clamped single-site families. (i) $H_{Q_R}(d)=H_{Q_C}(d)=2d+1$ for every $d$. (ii) If $m\neq0$
and $d\ge2$, then $N_R\not\subseteq N_C$ and $N_C\not\subseteq N_R$. (iii) If $m\neq0$, then $N_R\cap N_C=h\cdot
\mathcal P_{\le d-3}$ with $h=(a-a_0)(b-ma-c)(b-b_0)$, of dimension $(d-1)(d-2)/2$ for $d\ge3$ and $0$ for $d<3$.
\end{proposition}

Counts alone cannot rank these two \emph{single-site} designs on an exact slice; the proposition says nothing about
single-site versus joint evidence in a trained model. 

\emph{Proof.} (i) Count the constraints. Each reachable set comprises two lines, whose vanishing polynomials are
multiples of a degree-2 generator $g$. Thus $N=g\cdot\mathcal P_{\le d-2}$ has dimension $d(d-1)/2$ and profile
$\binom{d+2}{2}-d(d-1)/2=2d+1$.

(ii) Each generator witnesses incomparability. For $d\ge2$, $g_R\in N_R$ but $g_R\notin N_C$ because
$(b-b_0)\nmid(b-ma-c)$ when $m\neq0$; symmetrically $g_C\in N_C\setminus N_R$. For $d\le1$ both spaces are zero.

(iii) Find the shared blind differences. $p\in N_R\cap N_C$ iff $p=g_Ru=g_Cv$; cancelling $(a-a_0)$ gives
$(b-ma-c)u=(b-b_0)v$. The linear forms are non-associate when $m\neq0$. Unique factorisation in $\mathbb R[a,b]$
then gives $(b-b_0)\mid u$ and $p\in\langle h\rangle$; conversely, multiples of $h$ of degree at most $d$ lie in both spaces.
Equivalently, $N_R\cap N_C$ is generated by $\mathrm{lcm}(g_R,g_C)$ \citep{pistone1996groebner}. $\square$

\paragraph{Theorem~\ref{thm:operator}.} Find the vanishing polynomials on each arm, then intersect.
$I(A_0)=(a-a_0)$ because $A_0$ is an infinite subset of the line $a=a_0$.
$P$ vanishes on $\Gamma$, so $(P)\subseteq I(\Gamma)$. Conversely, if $Q$ vanishes on $\Gamma$ and $P\nmid Q$, then $P$
and $Q$ are coprime because $P$ is irreducible, so by B\'ezout's theorem $V(P)\cap V(Q)$ is finite, contradicting
$\Gamma\subseteq V(P)\cap V(Q)$ with $\Gamma$ infinite. Hence $I(\Gamma)=(P)$. Since $P$ is not an associate of
$a-a_0$, the two irreducibles are coprime and $I(A_0\cup\Gamma)=(a-a_0)\cap(P)=((a-a_0)P)$, whose nonzero elements
have degree at least $1+\deg P$. $\square$

Zariski density of $\Gamma$ in an irreducible $V(P)$ would not be enough. Over $\mathbb R$ it does
not give $I(\Gamma)=(P)$: for $P=(b-ma-c)^2$ the zero set is the irreducible line $b=ma+c$, but $I(\Gamma)=(b-ma-c)$,
so the onset is $2$, not $1+\deg P=3$. This is why the theorem asks for $P$ itself to be irreducible. The affine and
nondegenerate normalisation instances satisfy it; the transcendental instances are covered by the separate lemma below.

\paragraph{Irreducibility of the normalisation quartic.} To apply the operator theorem, rule out lower-degree
factors. Clearing the radical in
$e(a)=\beta+\omega a/\sqrt{\kappa a^2+\epsilon}$ gives $P(a,b)=(b-\beta)^2(\kappa a^2+\epsilon)-\omega^2a^2$, under
$\omega\neq0$, $\kappa>0$ and $\epsilon>0$ (for RMSNorm, $\epsilon$ is the model dimension times the normalisation
constant). As a quadratic in $X=b-\beta$ over $\mathbb R[a]$ it is $(\kappa a^2+\epsilon)X^2-\omega^2a^2$, with
discriminant $4\omega^2a^2(\kappa a^2+\epsilon)$. It factors over $\mathbb R[a]$ only if $\kappa a^2+\epsilon$ is a
square, which it is not, since it has two distinct complex roots; a factorisation with a factor free of $X$ would
need a common factor of $\kappa a^2+\epsilon$ and $\omega^2a^2$, and there is none. Hence $P$ is irreducible of degree
$\DstarRMSNormDegP$ and not an associate of $a-a_0$, and the recompute branch is an infinite subset of $V(P)$, so
Theorem~\ref{thm:operator} gives $d^*=\DstarRMSNorm$.

\paragraph{Transcendence of the analysed slices.}
The following lemma rules out exact polynomial blind differences on the stated softmax and gated slices.
\begin{lemma}
Let $e(a)=\omega\,\sigma(ga)$ with $\sigma(x)=(1+\exp(-x))^{-1}$ and real $\omega,g\neq0$. No nonzero $P\in\mathbb
R[a,b]$ satisfies $P(a,e(a))\equiv0$ on an interval with nonempty interior.
\end{lemma}
\emph{Proof.} A polynomial relation for the sigmoid would imply one for the exponential.
Suppose $P(a,e(a))\equiv0$ with $P\neq0$. Then $e$ is algebraic over $\mathbb R(a)$: it satisfies a polynomial
equation with rational-function coefficients. Since $e=\omega/(1+\exp(-ga))$, the rational expression
$E:=\exp(-ga)=\omega/e-1$ is also algebraic over $\mathbb R(a)$. Clearing denominators gives
$\sum_{j=0}^np_j(a)\exp(-jga)=0$ with $p_j\in\mathbb R[a]$; divide out any power of $E$ to take $p_0\neq0$.
This expression is real analytic, so vanishing on an interval with nonempty interior implies vanishing identically.
Now let $a\to+\infty$ if $g>0$, or $a\to-\infty$ if $g<0$. Terms with $j\ge1$ decay exponentially while $p_j$
grows at most polynomially, forcing $p_0(a)\to0$, impossible for a nonzero polynomial. $\square$

A two-key softmax edge $e(a)=\omega\exp(a)/(\exp(a)+\exp(\kappa))=\omega\,\sigma(a-\kappa)$ with $\omega\neq0$ has the
lemma's form after a shift, so $I(\Gamma)=(0)$ and $d^*=\DstarSoftmax$ on that slice. For a gated linear-attention edge
$e(a)=(wa)\,\sigma(ga)$ with $w,g\neq0$, a relation $P(a,e(a))\equiv0$ would make $\sigma(ga)=e/(wa)$ algebraic over
$\mathbb R(a)$ on a subinterval with nonempty interior avoiding $a=0$, and the lemma applies again. Both statements concern the analysed slice
in the chosen coordinates (Remark~\ref{rem:scope}).

\paragraph{Proposition~\ref{prop:ksite}.} On arm $i$ the swept coordinate $t$ parameterises the arm, and every
coordinate is a polynomial in $t$ of degree at most $\eta_i$ (a constant has degree $0\le\eta_i$). Restricting a
degree-$d$ hypothesis gives a univariate polynomial of degree at most $d\eta_i$, hence at most $d\eta_i+1$
independent constraints. The $k$ arms agree at the clean configuration, imposing $k-1$ constraints and reducing
$\sum_i(d\eta_i+1)$ to $1+d\sum_i\eta_i$. Subtracting the restriction rank from the class dimension gives the bound.

The degrees must include all downstream compositions: with edges $A\to B\to C$ of degree $p$, patching $A$ leaves $b$ of degree
$p$ but $c$ of degree $p^2$. On a chain with $k=\KCEk$ sites, edge degree $\KCEe$ and $d=\KCEd$, using the edge degree
for every arm gives $\KCEBound$, whereas the exact rank (over a finite field, generic coefficients, edges shared across
arms) is $\KCERankExact$; the composed bound, $\KCEComposedBound$, holds.

\emph{Affine arms attain the bound.} Translate the clean state to the origin. Under the feed-forward order of
the sites, the arm directions form a triangular matrix with unit diagonal, so an invertible linear map sends the
arms to the coordinate axes. On the axes, the restrictions of $1$ and $x_i^r$ ($1\le i\le k$, $1\le r\le d$) are
linearly independent, so the rank is exactly $1+kd$. A Cartesian product with at least $d+1$ distinct values per axis
determines every polynomial of degree at most $d$ in each variable, so for $d=3$ four values per axis suffice. At degree $3$, single-site evidence then leaves unresolved $\KBlindTwo\%$ of the
class at two sites, $\KBlindThree\%$ at three, $\KBlindFour\%$ at four and $\KBlindFive\%$ at five.

\paragraph{Numerical checks of the exact results.} The left panel of Figure~\ref{fig:regimes} corroborates the
operator-slice results. On each of three sampling spans, singular value decomposition (SVD) fits a candidate
polynomial, the right singular vector for the smallest singular value of the column-scaled design. Its relative
residual reaches numerical precision at $d^*$ on the affine and RMSNorm slices, while the plotted softmax residual
decreases over the tested degrees. Numerics cannot certify that no polynomial relation exists; for the
transcendental slices the proof is the lemma above.

% ===================================================================== C ====
\section{Glossary}
\label{app:glossary}

\begin{center}\small
\begin{tabularx}{\textwidth}{@{}l>{\raggedright\arraybackslash}X@{}}
\toprule
term & meaning \\
\midrule
chart & a map from the sites' outputs to low-dimensional coordinates $z=(a,b,\dots)$ \\
reachable set $R_Q$ & the chart states an intervention family produces \\
blind space $N(Q,\mathcal F)$ & hypothesis differences in the class $\mathcal F$ that vanish on all of $R_Q$ \\
resolving preorder $\succeq_{\mathcal F}$ & $Q_1\succeq_{\mathcal F}Q_2$ when $N(Q_1,\mathcal F)\subseteq N(Q_2,\mathcal F)$: $Q_1$ resolves at least what $Q_2$ resolves \\
onset $d^*$ & the least polynomial degree at which the blind space is nonzero \\
recompute / clamped generator & on an affine slice, the degree-2 generators of the respective single-site blind spaces: $g_R=(a-a_0)(b-ma-c)$ and $g_C=(a-a_0)(b-b_0)$ \\
energy $E_Q(h)$ & mean square of a hypothesis difference over the family's states, $\theta^\top G_Q\theta$ \\
held-out $J/S$ energy ratio $r(h)$ & $E_J(h)/E_S(h)$ on test rows, for $h$ defined from training rows only \\
S-weak & a direction with a large $r(h)$: much less constrained by single-site than joint evidence \\
edge strength $s$ & $\mathrm{sd}(b\mid\text{arm }A)/\mathrm{sd}(b\mid\text{arm }B)$ strong if $s\ge0.10$; computed on training single-site rows for the named direction, and on all single-site rows in the support-matching and IOI analyses \\
same-family envelope & the null band for optimised directions built by splitting the single-site family (\S\ref{sec:pitfalls}) \\
capture, cell, head pair & a capture is one (site pair, seed) run; a cell is a model $\times$ task. In \S\ref{sec:ioi}, a head pair is a (source head, Name Mover) pair, and a capture is that pair with a prompt \\
transport contrast $T_C$ & how much $B$'s measured effect changes when $A$ moves from its clean output to $u$: the four-point contrast $Y(u,v)-Y(u,0)-Y(0,v)+Y(0,0)$ of \S\ref{sec:ioi} \\
B-arm swing & the range of the readout over the single-site $B$ arm; the denominator of every ``\% of swing'' \\
working-model identification width & at a joint state, the largest $|T_C|$ inside a fit's nominal $95\%$ Wald region under the working model; summarised by a median over joint states within each capture, then a median over captures; neither an observed error nor a calibrated bound \\
design gap & the ratio, joint over single-site, of the smallest eigenvalue of the per-observation residualised design Gram of the cross block \\
\bottomrule
\end{tabularx}
\end{center}

% ===================================================================== D ====
\section{Protocol}
\label{app:protocol}

\paragraph{Protocol at a glance.} The following protocol underlies Sections~\ref{sec:practical} and~\ref{sec:pitfalls}; support and chart variants are
specified with each comparison. Profiles in the appendix are reported as $\log_{10}(E_S/E_J)=-\log_{10}r$, so the
\emph{S-weak end} (single-site evidence weaker) is at the bottom and the \emph{reverse end} (joint evidence weaker) at
the top; the main text's Figure~\ref{fig:regimes} plots the opposite sign.
\begin{itemize}[leftmargin=1.2em,itemsep=1pt,topsep=1pt,parsep=0pt]
\item \textbf{Units.} A \emph{capture} is one (site pair, seed) run, the seed selecting prompt and donors; a
\emph{cell} is a model $\times$ task. Table~\ref{tab:predict} has $\MatchedN$ float32 captures in $\PredCells$ cells and
$\HeldoutPrompts$ (cell, prompt) groups. Captures share models, tasks and often prompts, so counts are consistency
checks, and we make no population claim.
\item \textbf{Sites and chart.} Two attention-block outputs, each projected on the unembedding difference of the
task's two answer tokens, then whitened with statistics from training rows (the searched profiles use
all-row statistics, which is equivalent in exact arithmetic; see below). The head-level IOI analysis of
\S\ref{sec:ioi} uses a different chart.
\item \textbf{Single-site family $S$.} $32$ donors $\times$ $\NAlpha$ interpolation coefficients $\alpha\in[-0.5,1.5]$
on each arm: patch $A$ and recompute $B$, or patch $B$ with $A$ clean. Each row is one intervention, giving $448$ rows.
\item \textbf{Joint family $J$.} $256$ rows on the same $\alpha$ grid, pairing donors and coefficients in a fixed
sparse pattern; it is not a full product grid.
\item \textbf{Class and split.} Polynomials of total degree at most $3$ ($10$ dimensions). Rows are split in half
($S$: $224/224$; $J$: $128/128$); directions are named or fitted on training rows, frozen, and scored on test rows.
\item \textbf{Screen.} Captures with $\mathrm{sd}(Y_{\mathrm{joint}})<0.15$ are excluded. In
Sections~\ref{sec:practical} and~\ref{sec:pitfalls} this is the only use of readouts, and it makes the analysed sample
readout-conditioned.
\end{itemize}

\paragraph{Families.} Each capture patches two sites. A donor--coefficient pair sets a site's output to
$o_{\mathrm{clean}}+\alpha\,(o_{\mathrm{donor}}-o_{\mathrm{clean}})$, where $o_{\mathrm{donor}}$ is the site's output on
the donor prompt. The single-site family has, for each of $32$ donors and each
of the $\NAlpha$ coefficients $\alpha\in\{-0.5,-\tfrac16,\tfrac16,0.5,\tfrac56,\tfrac76,1.5\}$, one row patching $A$
with $B$ recomputed and one row patching $B$ with $A$ clean: $448$ rows. The support-matched joint family has $256$
rows. Row $r=16i+j$ ($i,j\in\{0,\dots,15\}$) pairs donors $2i$ and $2j$ with the coefficient indices $(i+j)\bmod 7$
and $(3i+j)\bmod 7$; on each axis it uses $112$ of the $224$ single-site (donor, coefficient) pairs, a sparse subset
of their Cartesian product. The query-chart captures of \S\ref{sec:ioi} index donors by donor id instead. Both joint
families of Appendix~\ref{app:matched} are collected in one pass with the single-site family, on identical prompts,
donors and weights.

\paragraph{Training statistics and searched profiles.} For the named direction, the edge fit, whitening and
column scaling use training rows only. The searched
polynomial profiles (Figure~\ref{fig:regimes}, right, \S\ref{sec:pitfalls} and Appendices~\ref{app:matched},
\ref{app:chart} and~\ref{app:three}) estimate whitening and column scaling on all analysed rows. For a fixed chart
and the complete, degree-ordered monomial bases used there, changing to training-row statistics is an invertible
lower-triangular change of feature basis with positive diagonal. This leaves the Cholesky-reduced covariant
eigenproblem unchanged in exact arithmetic. On identical row partitions, the fitted eigenfunctions therefore agree
up to scale, and the held-out profiles and same-family envelopes are unchanged. Numerically this holds only where the
held-out energies are resolved: at the clamped-weak end of Appendix~\ref{app:chart} the clamped family has an exact
polynomial kernel, so those energies are at rounding level under either choice. This does not cover chart selection
itself or the Euclidean-ridge checks of Appendices~\ref{app:robust} and~\ref{app:noninvariant}.

\paragraph{Precision.} All support-matched captures behind Table~\ref{tab:predict} and all IOI query-chart captures
are float32. The unmatched-protocol captures of GPT-2 on IOI and of Qwen3.5-9B-Base were produced with a bfloat16
model, so their recorded states as well as their readouts carry bfloat16 rounding; the other GPT-2 unmatched captures
are float32. Comparisons with those unmatched-protocol captures therefore also change precision.

\paragraph{Readout screen.} A capture is analysed only if $\mathrm{sd}(Y_{\mathrm{joint}})\ge0.15$. No readout enters
the naming or scoring of a direction, but the analysed sample is conditioned on the readout through this screen.
In the IOI query-chart analysis, $36$ published and $36$ control captures were designed and $\QCN$ and $\QCCtrlN$ kept;
the collector dropped captures that failed the screen or raised an exception without recording which.

\paragraph{Same-family envelope.} For optimised directions we compare the held-out profile with a same-family envelope.
We split $S$ into two stand-in families and run the identical train--freeze--score procedure. Within each cell, we
pool the other captures' null profiles and take the $2.5$th percentile of their minima and the $97.5$th percentile
of their maxima, leaving each capture's own nulls out. In the full-grid single-site--joint comparison, each stand-in
family has $112$ training and $112$ test rows, against $224/224$ for $S$ and $128/128$ for $J$. These sample sizes differ,
and captures share prompts, so the nominal $2.5\%$ escape rate is not established for the tested procedure. We report escape counts
descriptively and any $p$-values as nominal.

\paragraph{Applying the audit to another circuit.} (1) Declare $\mathcal F$; the chosen polynomial degree limits which
hypothesis differences the audit can assess. (2) Sweep the intervention family and record the internal states, not just
the readout. (3) Form $G_Q$ and read off its low-energy directions; with a second family, solve the covariant
pencil. (4) Before comparing two families, check that their reachable sets span the same region; for optimised
directions, compare with a same-family envelope, remembering that its $p$-values are nominal. (5) For a transport
claim, execute the clamped and joint corners and measure the numerical floor by re-executing them in a different
batch composition. Only steps (2) and (5) need GPU time.

% ===================================================================== E ====
\section{Support matching in full}
\label{app:matched}

The range comparison's registration and its results first entered version control together,
so their order cannot be verified. Two further protocols separate the effect of removing extrapolation from that of enlarging the joint family.
\emph{Interior} restricts the single-site family to $\alpha\in[0,1]$ and leaves the joint family at $\alpha=1$; it
needs no new data but is conservative rather than exact. \emph{Symmetric} restricts both families to the interior
coefficients, which answers the objection that matching helped only by enlarging the joint family. Across all
$\MatchedN$ support-matched captures the reverse end escapes the same-family envelope in $\EdgeFrozenEscapes$
captures under the unmatched protocol, $\EdgeInteriorEscapes$ under interior, $\EdgeSymmetricEscapes$ under symmetric
and $\EdgeMatchedEscapes$ under matched.

\begin{figure}[h]
\centering
\includegraphics[width=\textwidth]{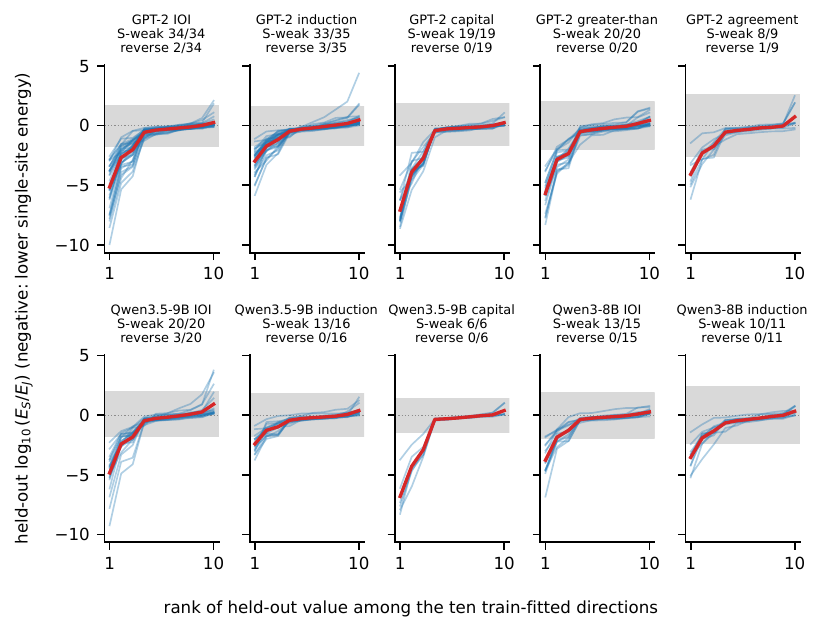}
\caption{Held-out profiles behind Table~\ref{tab:matched-full}, support-matched protocol (joint family on the single-site $\alpha$ grid, a sparse design). For each capture (thin
blue lines) the ten generalized eigen-directions are fitted on training rows and frozen; the plot shows their held-out
$\log_{10}(E_S/E_J)$ energy ratios \emph{sorted by held-out value} (order statistics). Negative values indicate lower held-out energy under single-site patching; Figure~\ref{fig:regimes} uses the opposite
sign, $\log_{10}r$.
The S-weak end is the minimum over the ten frozen directions, not the named generator of \S\ref{sec:predict}, and need
not be the first training eigenvector. The thick red line is the mean of the held-out order statistics. The grey band applies the percentiles of
Appendix~\ref{app:protocol} to the same-family nulls of all captures in the cell, pooled, and is shown for display; the
escape counts in the panel titles and in Table~\ref{tab:matched-full} leave each capture's own nulls out, as in
Appendix~\ref{app:protocol}. In GPT-2 $\times$ agreement, a reverse endpoint exceeds its leave-one-capture-out upper bound
while remaining below the displayed upper bound.}
\label{fig:spectrum}
\end{figure}

\begin{figure}[h]
\centering
\includegraphics[width=\textwidth]{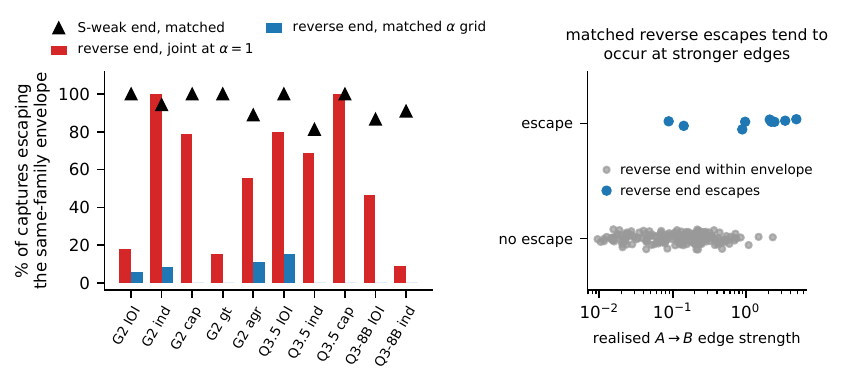}
\caption{Support matching and escapes of the same-family envelope. \emph{Left:} percentages of captures whose
optimised profiles escape at the S-weak or reverse (joint-weak) end: red bars, reverse end with the joint family at $\alpha=1$; blue bars, reverse end on
the matched grid; black triangles, S-weak end on the matched grid. Matching reduces reverse escapes.
\emph{Right:} each point is a capture; edge strength (the matched analysis's all-rows statistic, logarithmic axis)
is plotted against the binary matched-protocol reverse-escape indicator, with vertical jitter. Blue points escape at the reverse end;
grey points do not escape at that end, although their S-weak end may escape. Escapes are associated with stronger edges. Tick labels: G2, GPT-2; Q3.5, Qwen3.5-9B-Base; Q3-8B, Qwen3-8B;
ind, induction; cap, capital; gt, greater-than; agr, agreement.}
\label{fig:matched}
\end{figure}

Edge strength is associated with a reverse escape under the matched (Spearman $\EdgeMatchedRho$), interior
($\EdgeInteriorRho$) and symmetric ($\EdgeSymmetricRho$) protocols, and only weakly under the unmatched one
($\EdgeFrozenRho$). Under the matched protocol the median edge strength is $\EdgeMatchedMedEscaped$ for escapes against
$\EdgeMatchedMedNot$ for non-escapes. These correlations, the edge-strength medians and survivor count of \S\ref{sec:pitfalls}, and
Figure~\ref{fig:matched} use the edge statistic stored with the matched analysis: the arm-$A$/arm-$B$ spread ratio
computed on all single-site rows rather than on training rows only; the paragraph on the realised edge below uses the
training-row statistic $s$ of \S\ref{sec:predict}. Figure~\ref{fig:spectrum} shows the held-out profiles behind
Table~\ref{tab:matched-full}, and Figure~\ref{fig:matched} the escape rates and their association with edge strength.
Table~\ref{tab:predict-unmatched} repeats the $g_R$ columns of Table~\ref{tab:predict} against the unmatched joint
family on the unmatched-protocol captures.

\begin{table}[!htbp]
\centering\small
\begin{tabular}{llcccc}
\toprule
model $\times$ task & protocol & S-weak end & $q$ & reverse end & $q$ \\
\midrule
\multirow{4}{*}{GPT-2 $\times$ IOI}
 & joint at $1$ & \MGPTIOIFrozenSblind & $\MGPTIOIFrozenQS$ & \MGPTIOIFrozenJblind & $\MGPTIOIFrozenQJ$ \\
 & interior & \MGPTIOIInteriorSblind & $\MGPTIOIInteriorQS$ & \MGPTIOIInteriorJblind & $\MGPTIOIInteriorQJ$ \\
 & symmetric & \MGPTIOISymmetricSblind & $\MGPTIOISymmetricQS$ & \MGPTIOISymmetricJblind & $\MGPTIOISymmetricQJ$ \\
 & \textbf{matched} & \MGPTIOIMatchedSblind & $\MGPTIOIMatchedQS$ & \MGPTIOIMatchedJblind & $\MGPTIOIMatchedQJ$ \\
\midrule
\multirow{4}{*}{GPT-2 $\times$ induction}
 & joint at $1$ & \MGPTIndFrozenSblind & $\MGPTIndFrozenQS$ & \MGPTIndFrozenJblind & $\MGPTIndFrozenQJ$ \\
 & interior & \MGPTIndInteriorSblind & $\MGPTIndInteriorQS$ & \MGPTIndInteriorJblind & $\MGPTIndInteriorQJ$ \\
 & symmetric & \MGPTIndSymmetricSblind & $\MGPTIndSymmetricQS$ & \MGPTIndSymmetricJblind & $\MGPTIndSymmetricQJ$ \\
 & \textbf{matched} & \MGPTIndMatchedSblind & $\MGPTIndMatchedQS$ & \MGPTIndMatchedJblind & $\MGPTIndMatchedQJ$ \\
\midrule
\multirow{4}{*}{GPT-2 $\times$ capital}
 & joint at $1$ & \MGPTCapFrozenSblind & $\MGPTCapFrozenQS$ & \MGPTCapFrozenJblind & $\MGPTCapFrozenQJ$ \\
 & interior & \MGPTCapInteriorSblind & $\MGPTCapInteriorQS$ & \MGPTCapInteriorJblind & $\MGPTCapInteriorQJ$ \\
 & symmetric & \MGPTCapSymmetricSblind & $\MGPTCapSymmetricQS$ & \MGPTCapSymmetricJblind & $\MGPTCapSymmetricQJ$ \\
 & \textbf{matched} & \MGPTCapMatchedSblind & $\MGPTCapMatchedQS$ & \MGPTCapMatchedJblind & $\MGPTCapMatchedQJ$ \\
\midrule
\multirow{4}{*}{GPT-2 $\times$ greater-than}
 & joint at $1$ & \MGPTGtFrozenSblind & $\MGPTGtFrozenQS$ & \MGPTGtFrozenJblind & $\MGPTGtFrozenQJ$ \\
 & interior & \MGPTGtInteriorSblind & $\MGPTGtInteriorQS$ & \MGPTGtInteriorJblind & $\MGPTGtInteriorQJ$ \\
 & symmetric & \MGPTGtSymmetricSblind & $\MGPTGtSymmetricQS$ & \MGPTGtSymmetricJblind & $\MGPTGtSymmetricQJ$ \\
 & \textbf{matched} & \MGPTGtMatchedSblind & $\MGPTGtMatchedQS$ & \MGPTGtMatchedJblind & $\MGPTGtMatchedQJ$ \\
\midrule
\multirow{4}{*}{GPT-2 $\times$ agreement}
 & joint at $1$ & \MGPTAgrFrozenSblind & $\MGPTAgrFrozenQS$ & \MGPTAgrFrozenJblind & $\MGPTAgrFrozenQJ$ \\
 & interior & \MGPTAgrInteriorSblind & $\MGPTAgrInteriorQS$ & \MGPTAgrInteriorJblind & $\MGPTAgrInteriorQJ$ \\
 & symmetric & \MGPTAgrSymmetricSblind & $\MGPTAgrSymmetricQS$ & \MGPTAgrSymmetricJblind & $\MGPTAgrSymmetricQJ$ \\
 & \textbf{matched} & \MGPTAgrMatchedSblind & $\MGPTAgrMatchedQS$ & \MGPTAgrMatchedJblind & $\MGPTAgrMatchedQJ$ \\
\midrule
\multirow{4}{*}{Qwen3.5-9B $\times$ IOI}
 & joint at $1$ & \MQwenIOIFrozenSblind & $\MQwenIOIFrozenQS$ & \MQwenIOIFrozenJblind & $\MQwenIOIFrozenQJ$ \\
 & interior & \MQwenIOIInteriorSblind & $\MQwenIOIInteriorQS$ & \MQwenIOIInteriorJblind & $\MQwenIOIInteriorQJ$ \\
 & symmetric & \MQwenIOISymmetricSblind & $\MQwenIOISymmetricQS$ & \MQwenIOISymmetricJblind & $\MQwenIOISymmetricQJ$ \\
 & \textbf{matched} & \MQwenIOIMatchedSblind & $\MQwenIOIMatchedQS$ & \MQwenIOIMatchedJblind & $\MQwenIOIMatchedQJ$ \\
\midrule
\multirow{4}{*}{Qwen3.5-9B $\times$ induction}
 & joint at $1$ & \MQwenIndFrozenSblind & $\MQwenIndFrozenQS$ & \MQwenIndFrozenJblind & $\MQwenIndFrozenQJ$ \\
 & interior & \MQwenIndInteriorSblind & $\MQwenIndInteriorQS$ & \MQwenIndInteriorJblind & $\MQwenIndInteriorQJ$ \\
 & symmetric & \MQwenIndSymmetricSblind & $\MQwenIndSymmetricQS$ & \MQwenIndSymmetricJblind & $\MQwenIndSymmetricQJ$ \\
 & \textbf{matched} & \MQwenIndMatchedSblind & $\MQwenIndMatchedQS$ & \MQwenIndMatchedJblind & $\MQwenIndMatchedQJ$ \\
\midrule
\multirow{4}{*}{Qwen3.5-9B $\times$ capital}
 & joint at $1$ & \MQwenCapFrozenSblind & $\MQwenCapFrozenQS$ & \MQwenCapFrozenJblind & $\MQwenCapFrozenQJ$ \\
 & interior & \MQwenCapInteriorSblind & $\MQwenCapInteriorQS$ & \MQwenCapInteriorJblind & $\MQwenCapInteriorQJ$ \\
 & symmetric & \MQwenCapSymmetricSblind & $\MQwenCapSymmetricQS$ & \MQwenCapSymmetricJblind & $\MQwenCapSymmetricQJ$ \\
 & \textbf{matched} & \MQwenCapMatchedSblind & $\MQwenCapMatchedQS$ & \MQwenCapMatchedJblind & $\MQwenCapMatchedQJ$ \\
\midrule
\multirow{4}{*}{Qwen3-8B $\times$ IOI}
 & joint at $1$ & \MQIOIFrozenSblind & $\MQIOIFrozenQS$ & \MQIOIFrozenJblind & $\MQIOIFrozenQJ$ \\
 & interior & \MQIOIInteriorSblind & $\MQIOIInteriorQS$ & \MQIOIInteriorJblind & $\MQIOIInteriorQJ$ \\
 & symmetric & \MQIOISymmetricSblind & $\MQIOISymmetricQS$ & \MQIOISymmetricJblind & $\MQIOISymmetricQJ$ \\
 & \textbf{matched} & \MQIOIMatchedSblind & $\MQIOIMatchedQS$ & \MQIOIMatchedJblind & $\MQIOIMatchedQJ$ \\
\midrule
\multirow{4}{*}{Qwen3-8B $\times$ induction}
 & joint at $1$ & \MQIndFrozenSblind & $\MQIndFrozenQS$ & \MQIndFrozenJblind & $\MQIndFrozenQJ$ \\
 & interior & \MQIndInteriorSblind & $\MQIndInteriorQS$ & \MQIndInteriorJblind & $\MQIndInteriorQJ$ \\
 & symmetric & \MQIndSymmetricSblind & $\MQIndSymmetricQS$ & \MQIndSymmetricJblind & $\MQIndSymmetricQJ$ \\
 & \textbf{matched} & \MQIndMatchedSblind & $\MQIndMatchedQS$ & \MQIndMatchedJblind & $\MQIndMatchedQJ$ \\
\bottomrule
\end{tabular}
\caption{Escapes of the same-family envelope by the optimised profile's S-weak end and reverse end, by protocol, over
all $\MatchedN$ support-matched captures; denominators are the screened captures in each cell. $q$ is a
Holm-adjusted binomial $p$-value against the envelope's nominal $2.5\%$ escape rate, adjusted across cells for each
protocol and end; it is nominal (\S\ref{sec:pitfalls}). Under the matched protocol no cell retains a nominally
significant reverse end; under the interior protocol GPT-2 $\times$ induction and Qwen3.5-9B $\times$ induction do,
and under the symmetric protocol GPT-2 $\times$ induction does.}
\label{tab:matched-full}
\end{table}

\begin{table}[!htbp]
\centering\small
\begin{tabular}{lcc}
\toprule
model $\times$ task & median $r(g_R)$ & $r(g_R)>10\times$ \\
\midrule
GPT-2 $\times$ IOI          & $\PredGPTIOIRatio$  & \PredGPTIOICount \\
GPT-2 $\times$ induction    & $\PredIndGPTRatio$  & \PredIndGPTCount \\
GPT-2 $\times$ capital      & $\PredCapGPTRatio$  & \PredCapGPTCount \\
GPT-2 $\times$ greater-than & $\PredGtGPTRatio$   & \PredGtGPTCount \\
GPT-2 $\times$ agreement    & $\PredAgrGPTRatio$  & \PredAgrGPTCount \\
Qwen3.5-9B $\times$ IOI     & $\PredQwenIOIRatio$ & \PredQwenIOICount \\
Qwen3.5-9B $\times$ induction & $\PredIndQwenRatio$ & \PredIndQwenCount \\
Qwen3.5-9B $\times$ capital & $\PredCapQwenRatio$ & \PredCapQwenCount \\
\midrule
all                         &                     & \PredTotalCount \\
\bottomrule
\end{tabular}
\caption{Held-out $J/S$ energy ratios of the named generator $g_R$ against the unmatched joint family (raw donors,
$\alpha=1$), repeating its columns of Table~\ref{tab:predict} on the unmatched-protocol captures. The qualitative result persists ($\PredTotalCount$ above $10\times$); the magnitude of the $J/S$
ratio depends on the joint support, and the cohorts differ in precision by cell (Appendix~\ref{app:protocol}).}
\label{tab:predict-unmatched}
\end{table}

\paragraph{The named generator and the one-at-a-time baseline.} In the captures, writing $b=ma+c+\varepsilon$ with the recorded
clean $b_0$, the clamped generator equals $m(a-a_0)^2+(a-a_0)(ma_0+c-b_0+\varepsilon)$. Both generators clear $10\times$ in
$\OfatBoth$ captures, $g_R$ alone in $\OfatOnlyRecompute$, $g_C$ alone in $\OfatOnlyClampedCount$, and neither in
$\OfatNeither$, a net difference of $\OfatNetDifference$. The median $r(g_R)$ is $\EdgeBlindWeak$ on weak-edge captures and
$\EdgeBlindStrong$ on the $\EdgeNStrong$ strong-edge ones, of which $\EdgeStrongGtTen$ exceed $10\times$. Of the surviving reverse
escapes of \S\ref{sec:pitfalls}, $\ReverseSurvivorsWeak$ fall below the strong-edge threshold. For the GPT-2 IOI captures, the
single-site range of $a$ exceeds its range under the unmatched joint family by a median factor of $\FigOneBCaptureMedian$,
the support gap behind Figure~\ref{fig:reachable}. Fitting the edge helps most where it is steep: the within-capture ratio
$r(g_R)/r(g_C)$ rises with the absolute whitened edge slope (Spearman $\OfatSpearman$).

\paragraph{The realised edge, measured.} On all $\EdgeN$ support-matched captures, the absolute Euclidean cosine between the
recompute and clamped generators, in the fixed whitened, column-scaled coefficient basis, falls as the training-row edge strength
$s$ (Appendix~\ref{app:glossary}) rises (Spearman $\EdgeCosRho$; with the absolute whitened slope it is $\EdgeCosRhoSlope$). Its median is
$\EdgeWeakCos$ below $s=0.10$ and $\EdgeStrongCos$ on the $\EdgeNStrong$ captures at or above it. These cosines are
parameterisation-dependent and descriptive; they do not show that the two designs are equivalent. On the
strong-edge captures the median held-out ratio is $\EdgeStrongSepGC$ for $g_C$ and $\EdgeBlindStrong$ for $g_R$, and
the median within-capture ratio $r(g_R)/r(g_C)$ is $\EdgeStrongRatio$. The design comparison rests on these energy
ratios and on Table~\ref{tab:predict}: the one-at-a-time generator accounts for most threshold exceedances in this
chart, and the recompute generator has an additional advantage on stronger edges.

% ===================================================================== F ====
\section{Chart selection in full}
\label{app:chart}

This comparison has no dated preregistration entry. This experiment compares recompute single-site evidence $R$ with clamped single-site evidence $C$, not single-site
with joint evidence. In the unembedding chart the absolute whitened edge slope is small (median $\RCUchartChartM$), so the
two designs almost coincide and both ends of their optimised profiles are nominally significant in only
$\RCUchartMutual$ cells. We therefore chose the
downstream coordinate by maximising a regularised variance ratio between the arms, using states only; the corresponding
spread ratio is $\mathrm{sd}(b\mid\text{arm }A)/\mathrm{sd}(b\mid\text{arm }B)$. In that chart the edge is strong (median $|m|=\RCChartM$). Scoring the held-out $R/C$ energy profile of the
train-fitted generalized eigen-directions, the recompute-weak end escapes the same-family envelope in $\RCPooledR$
captures, and the clamped-weak end escapes in every retained capture; both ends are nominally significant in
$\RCMutual$ cells.

The chart was fitted on states of the capture on which it was scored. In addition to the readout screen,
this analysis excludes captures with an arm-$A$/arm-$B$ spread ratio above $50$: $\RCDropped$ captures in the
own-chart and donor-holdout analyses. Holding out donors is not enough, because the chart is still fitted on the
capture's own prompt; the recompute-weak end still escapes in $\RCHoldoutPooledR$ captures.

With the chart of a different seed of the same layer pair, the
recompute-weak end escapes in $\RCXseedPooledR$ captures and both ends are nominally significant in $\RCXseedMutual$
cells (GPT-2 capital, GPT-2 IOI and Qwen3.5-9B capital). The cross-seed analysis covers the eight cells with an
available source chart, and its artifact predates the Qwen3-8B cross-seed captures, so the cohorts differ and the
comparison is not a paired effect estimate. For example, GPT-2 induction falls from $\RCGPTIndRblind$ to
$\RCGPTIndXseedRblind$.

A seed change does not partition content either:
$\AuditXseedSamePrompt$ evaluated captures share the chart source's base prompt, and donors are drawn from a shared
pool without a cross-seed exclusion rule. This is a cross-seed transfer control, not a disjoint-content
validation. The reversal between cells under this control is consistent with selection; the differing cohorts limit its interpretation.

The selection criterion is a variance ratio whose denominator is estimated from at most $16$ donor directions in a
much larger space ($768$ dimensions for GPT-2). In the GPT-2 captures for which this was logged, the selected direction had
negligible projection onto the span of the donor directions used to estimate the denominator, so its variance ratio was controlled by the
regulariser. The chart is therefore not ``the direction in which $A$ drives $B$'' but ``the direction in which $A$
drives $B$ and these particular donors do not''.

% ===================================================================== G ====
\section{Hypotheses that did not survive}
\label{app:negatives}

Two planned hypotheses had dated preregistration records and failed their criteria.

\paragraph{(1) Failed hypothesis: a circuit with high in-family faithfulness degrades reciprocally across families.} We fitted and froze
an edge-attribution circuit under each intervention family and scored it under both. The registered gate required
in-family faithfulness of at least $0.85$ and a cross-family drop of at least $0.25$ in both directions. It failed.
The recovered component sets largely overlapped, but overlap alone does not show that transfer cannot degrade. A
circuit is a set of components, whereas the ambiguity studied here lives in functional form, which a set cannot
express.

\paragraph{(2) Failed hypothesis: interpretability representations have distinctive weakly constrained subspaces.} We expected a logit
lens or a tuned lens to show S-weak subspaces distinguishable from matched compression controls. The registered success criterion required an effect outside the matched same-family and
compression nulls, a nontrivial held-out effect, and a consistent direction across multiple model-by-task units.
For $12$ GPT-2 captures in a four-dimensional chart, entries below are capture medians of the held-out energy share
$v_S=E_S/(E_S+E_J)$ of the least-energy training direction. The same-family null is rebuilt at each degree; each null
entry is the capture median of the per-capture $5$th percentile over five same-family resamples at the degree
indicated by its column header.

The registered hypothesis named a logit lens and a tuned lens; the arms that implement them differ from the standard
tools. The arm we call the attention-output unembedding maps each
last-token attention-module output $A$ to $\rho(A)=\bigl((A-\bar A)/(\operatorname{sd}(A)+10^{-6})\bigr)W_U^{\top}$,
evaluated through its isometric SVD reduction; it uses this custom standardisation rather than the model's learned final
LayerNorm and acts on attention outputs, not residual-stream states, so it is not the conventional logit lens. The arm
registered as the tuned lens is a least-squares linear surrogate fitted to this attention-output
unembedding of the first capture's patch-$A$-arm rows and reused across captures and
sites. Of its calibration rows, $\TLCalibOverlap$ fall in that capture's scored half; only the subsequent chart
fit is restricted to training rows.

\begin{center}\small
\begin{tabular}{lrrrr}
\toprule
 & \multicolumn{2}{c}{share $v_S$} & \multicolumn{2}{c}{same-family null ($5$th pct.)} \\
\cmidrule(lr){2-3}\cmidrule(lr){4-5}
representation & $\mathcal F=\mathcal P_{\le1}$ & $\mathcal F=\mathcal P_{\le2}$ & $\mathcal F=\mathcal P_{\le1}$ & $\mathcal F=\mathcal P_{\le2}$ \\
\midrule
raw activations & $\MCRawDegOne$ & $\MCRawDegTwo$ & $\MCRawNull$ & $\MCRawNullTwo$ \\
attention-output unembedding & $\MCLLDegOne$  & $\MCLLDegTwo$  & $\MCLLNull$ & $\MCLLNullTwo$ \\
linear surrogate of it & $\MCTLDegOne$  & $\MCTLDegTwo$  & $\MCTLNull$ & $\MCTLNullTwo$ \\
\midrule
random projection (matched control) & $\MCRandDegOne$ & $\MCRandDegTwo$ & $\MCRandNull$ & $\MCRandNullTwo$ \\
\bottomrule
\end{tabular}
\end{center}

\noindent At degree $1$ each row's capture-median share lies below its degree-1 null by a factor of
$\MCNullRatioOneLo$--$\MCNullRatioOneHi$; at degree $2$ it lies below its degree-2 null by a factor of
$\MCNullRatioTwoLo$--$\MCNullRatioTwoHi$. These factors are ratios of the capture-median null to the capture-median
share, not medians of within-capture ratios. Degree-3 values fall below the numerical resolution of the computation,
where even an orthogonal reparameterisation moves them substantially, so we do not report them.

The captures cover one model-by-task unit (GPT-2, IOI). The compression control is one fixed pair of random site projections reused across
captures rather than a distribution over projections. The captures therefore cannot establish the criterion, which
was not met; this shows neither that representation effects are absent nor that any representation separates. At degree $2$
the attention-output-unembedding share is below the random-projection share in $\MCLLBelowRandTwo$ captures, the raw-activation share in
$\MCRawBelowRandTwo$ and the surrogate's share in $\MCTLBelowRandTwo$; the direction is therefore not specific to the
attention-output unembedding, and these are descriptive comparisons within that one unit. The registration treats cross-representation
eigenvalue magnitudes as class-relative, so between-row differences in the table do not by themselves establish the
registered effect. A sparse-autoencoder arm was omitted by protocol, since no fixed released autoencoder was available
for these models. These results concern the implemented representations and do not establish a general conclusion
about tuned lenses.

% ===================================================================== H ====
\section{Per-cell results and stricter checks}
\label{app:units}

\paragraph{Per-cell results.} Table~\ref{tab:predict} gives, for each model $\times$ task cell, the median held-out $J/S$
energy ratio $r(g_R)$ of the recompute generator $g_R$, the number of captures in which that ratio exceeds $10\times$,
and the same count for the one-at-a-time (clamped) generator $g_C=(a-a_0)(b-b_0)$. Both generators are scored on the same
recompute single-site and joint states, and a large ratio means the direction is S-weak. The $10\times$ threshold is
descriptive. GPT-2 is GPT-2 small; Qwen3.5-9B is Qwen3.5-9B-Base.

\begin{table}[!htbp]
\centering\small
\caption{Held-out $J/S$ energy ratios of the geometry-named direction $g_R$, with the one-at-a-time generator $g_C$ as
baseline, by model $\times$ task cell. Both generators are scored on the same recompute single-site and joint states.
The ratio column gives the median $r(g_R)$; the other columns count captures above $10\times$.}
\label{tab:predict}
\begin{tabular}{lccc}
\toprule
model $\times$ task & median $r(g_R)$ & $r(g_R)>10\times$ & $r(g_C)>10\times$ \\
\midrule
GPT-2 $\times$ IOI                & $\PredGPTIOIMRatio$  & \PredGPTIOIMCount  & \PredGPTIOIMOfat \\
GPT-2 $\times$ induction          & $\PredIndGPTMRatio$  & \PredIndGPTMCount  & \PredIndGPTMOfat \\
GPT-2 $\times$ capital            & $\PredCapGPTMRatio$  & \PredCapGPTMCount  & \PredCapGPTMOfat \\
GPT-2 $\times$ greater-than       & $\PredGtGPTMRatio$   & \PredGtGPTMCount   & \PredGtGPTMOfat \\
GPT-2 $\times$ agreement          & $\PredAgrGPTMRatio$  & \PredAgrGPTMCount  & \PredAgrGPTMOfat \\
Qwen3.5-9B $\times$ IOI           & $\PredQwenIOIMRatio$ & \PredQwenIOIMCount & \PredQwenIOIMOfat \\
Qwen3.5-9B $\times$ induction     & $\PredIndQwenMRatio$ & \PredIndQwenMCount & \PredIndQwenMOfat \\
Qwen3.5-9B $\times$ capital       & $\PredCapQwenMRatio$ & \PredCapQwenMCount & \PredCapQwenMOfat \\
Qwen3-8B $\times$ IOI             & $\PredQIOIMRatio$    & \PredQIOIMCount    & \PredQIOIMOfat \\
Qwen3-8B $\times$ induction       & $\PredQIndMRatio$    & \PredQIndMCount    & \PredQIndMOfat \\
\midrule
all captures                      &                      & \PredMTotalCount   & \OfatRowCount \\
\bottomrule
\end{tabular}
\end{table}

\paragraph{Donor indices.} The row split in \S\ref{sec:predict} places states from the same donor on both
sides, since each donor contributes $\NAlpha$ coefficients. Under a common partition of donor indices that scores a joint row only if both of its
donor indices are held out, $g_R$ exceeds $10\times$ in $\DonorNamedCount$ captures and $g_C$ in $\OfatDonorCount$
(both $\OfatDonorBoth$, only $g_R$ $\OfatDonorOnlyRecompute$, only $g_C$ $\OfatDonorOnlyClampedCount$, neither
$\OfatDonorNeither$). This partition holds out donor indices, not donor content: in $\AuditDonorDupCaptures$ captures a
held-out donor's patch-$A$ trajectory is identical to a training donor's, and whether the donor prompts coincide was
not checked. It is therefore not a disjoint-content validation.

\paragraph{Prompt groups.} The $\MatchedN$ captures rest on $\HeldoutPrompts$ (cell, prompt) groups that share clean
activations, donor pools and task instances, because a capture's seed selects its prompt. Counting a group only if \emph{every} capture in it exceeds
$10\times$ under the row split, $g_R$ succeeds in $\HeldoutPromptStrict$ groups and $g_C$ in $\OfatPromptStrict$.
Resampling the groups with replacement, the $2.5$th percentile of the pooled fraction of captures with
$r(g_R)>10$ under the row split is $\HeldoutPromptBootLo\%$. Groups still share models and tasks.

\paragraph{Random directions.} For $200$ Gaussian directions per capture in the training-scaled coefficient basis, the
cell medians of the capture-specific $95$th percentiles range from $\HeldoutNullLo$ to $\HeldoutNullHi$, and $g_R$
exceeds its own capture's $95$th percentile in $\HeldoutBeatsNull$ captures.

% ===================================================================== I ====
\section{Three sites in a trained model}
\label{app:three}

\paragraph{Design.} Proposition~\ref{prop:ksite} counts rank on the population reachable set. Here we extend the
trained-model energy comparison to three GPT-2 attention blocks: three single-site arms, each site patched alone across the same
coefficient grid with the other two left to the network, against a support-matched joint family over donor triples,
in float32, with the same chart, split and envelope. The layer triples are $5$--$7$--$9$, $5$--$8$--$11$,
$6$--$9$--$11$ and $7$--$9$--$11$, with three seeds each.

\paragraph{Full rank, weakly constrained directions.} The design matrices have full numerical rank (singular-value
tolerance $10^{-9}$) under both families, so the comparison concerns ill-conditioning. The profile reports
single-site/joint energy, so more negative log ratios mean weaker single-site evidence; this is the opposite sign to
$\log_{10}r$ in the main text. For each capture we score every
frozen training eigenvector and take the minimum held-out log energy ratio $\log_{10}(E_S^{\mathrm{te}}/E_J^{\mathrm{te}})$. In the
$\ThreeDegThreeDim$-dimensional degree-3 class the median over captures of these minima is $-\ThreeDegThreeLogRatio$;
the profile minimum escapes the same-family envelope in $\ThreeDegThreeSblind$ captures and the profile maximum in
$\ThreeDegThreeReverse$. At degree $2$ ($\ThreeDegTwoDim$ dimensions) the corresponding median is
$-\ThreeDegTwoLogRatio$, and the profile minimum escapes in $\ThreeDegTwoSblind$.

% ===================================================================== J ====
\section{Further robustness checks and a synthetic check of the estimator}
\label{app:robust}

\paragraph{Charts and classes.} These are exploratory checks on the unmatched captures. They use a
Euclidean ridge $G_Q+10^{-10}I$ rather than the covariant pencil, and whitening and feature scaling estimated on all
rows; the data-derived charts and the non-polynomial classes therefore also use evaluation states. Only the
eigensolve is fitted on the training half, so these are not held-out validations in the sense of
\S\ref{sec:practical}. Both tables report the median evaluation-half $J/S$ ratio of the least-energy direction.

\begin{center}\small
\begin{tabular}{lr}
\toprule
chart (eight GPT-2 captures) & median $J/S$ ratio (Euclidean ridge) \\
\midrule
$u$ (unembedding difference) & $\CoordUSblind$ \\
top activation principal component & $\CoordPConeSblind$ \\
pooled intervention mean $-$ clean & $\CoordDiffSblind$ \\
four-dimensional chart (two components per site; degree $2$) & $\CoordPCtwodSblind$ \\
\bottomrule
\end{tabular}
\end{center}

\begin{center}\small
\begin{tabular}{lrr}
\toprule
class & GPT-2 ($12$ captures) & Qwen3.5-9B ($15$ captures) \\
\midrule
polynomial & $\BGPTPolySblind$ & $\BQwenPolySblind$ \\
fixed-grid radial basis (RBF; different class) & $\BGPTRbfSblind$ & $\BQwenRbfSblind$ \\
random Fourier (different class) & $\BGPTFourierSblind$ & $\BQwenFourierSblind$ \\
\bottomrule
\end{tabular}
\end{center}

\noindent A Taylor class with a zero centre has the identical design matrix to the polynomial row, so it is not an
independent check and is not reported. Magnitudes differ across classes, as Definition~\ref{def:blind} predicts, and
we never compare them as though the classes shared a geometry.

\paragraph{A synthetic check of subspace membership.} In synthetic union-of-slices data with a planted generator $g$,
the pencil estimator's least-energy direction lies close to the known degree-3 blind subspace
$\operatorname{span}\{g,ag,bg\}$ in a fixed coefficient basis: its Euclidean projection residual is $\PCResidZero$ at zero noise and at most
$\PCResidWorst$ up to noise $10^{-1}$, against a random-vector $1$st percentile of at least $\PCRandPOne$. This checks
subspace membership, not recovery of a unique generator, and it is basis-dependent.

% ===================================================================== K ====
\section{Non-invariant checks and worst-case bounds}
\label{app:noninvariant}

\paragraph{Covariant pencil.} Under $\varphi\mapsto L\varphi$ every term of the pencil of \S\ref{sec:pitfalls} transforms as
$L(\cdot)L^\top$, so its spectrum is invariant in exact arithmetic. In a numerical check of the unregularised pencil on
the unmatched-protocol captures, the largest cell median of the per-capture maximum relative eigenvalue change under a
reparameterisation is $\ReparamInvariance$. A Euclidean ridge on $G_Q$ and Euclidean coefficient angles or cosines are
not invariant; they are used only in Appendices~\ref{app:matched}, \ref{app:robust} and~\ref{app:noninvariant}, as
descriptive, Euclidean-ridge or non-invariance checks.

\paragraph{Euclidean ridge.} $G_Q+\lambda I$ does not transform covariantly under $\varphi\mapsto L\varphi$ for
non-orthogonal $L$. Under a reparameterisation of condition number $50$ on the Qwen3.5-9B IOI captures, the
$\lambda=10^{-6}$ spectrum moves by a median relative $\EucRegInvariance$, against $\CovRegInvariance$ for the covariant
regulariser of \S\ref{sec:pitfalls}.

\paragraph{Principal angles.} Euclidean angles between coefficient subspaces depend on the parameterisation of the
coefficient space. On the unmatched-protocol captures, the principal angles between the low-energy eigenspaces of the single-site and
joint training Grams have a median of $\AngleBefore^\circ$. Recomputing both eigenspaces after a random
reparameterisation of the same condition number gives $\AngleAfter^\circ$. Both summaries are medians over model--task
cells of within-cell medians of per-capture median angles. Because of this dependence we report no principal-angle quantity as evidence.

\paragraph{Worst-case spectral bounds are vacuous here.} Treating the finite intervention grid as a population, a
matrix Bernstein inequality \citep{tropp2015introduction} bounds $\|\widehat G_Q-G_Q\|$; for a two-sided deviation of a
Hermitian sum the intrinsic-dimension version carries a factor $8\,\mathrm{intdim}$ and holds only above a threshold
deviation. The Davis--Kahan bound \citep{yu2015davis} then limits $\|\sin\Theta\|_F$ for the bottom-$k$
eigenspace ($k=3$) by $2\sqrt k\,\|\widehat G-G\|/\gamma_k$, where $\gamma_k=\lambda_{k+1}-\lambda_k$ in increasing
order. For the joint family the median relative gap at that boundary is $\FSGPTEigengapBottomRel$ (GPT-2) and
$\FSQwenEigengapBottomRel$ (Qwen3.5-9B). With these gaps the bound exceeds its trivial value, so we report no numerical bound. The measured log--log concentration slopes of the single-site family, $\FSGPTRateSlope$ and
$\FSQwenRateSlope$ against a theoretical $-0.5$, are consistent with the rate but span under one decade of $n$.
Nothing here is used as a guarantee.

\paragraph{Held-out energies are not certificates.} For a direction frozen on training rows,
$v^\top\widehat G_{\mathrm{test}}v$ is an out-of-fit mean over the complementary rows and needs no eigengap.
The usual certificate conditions on training rows and applies a fixed-vector concentration inequality to an
independent test sample. It is not directly available here: test rows are the deterministic complement of training
rows in a fixed finite grid. We make no certification claim.

% ===================================================================== L ====
\section{IOI analysis in detail}
\label{app:ioi}

\begin{table}[t]
\centering\small
\caption{Fit, prediction, Wald envelopes and coverage of the working model on the published IOI head pairs
(exploratory; three base prompts; $\MisN$ captures). Compare the prediction errors with predicting zero, and the envelope widths with their coverage. $R^2$ rows: median over ten donor partitions of the median over captures; ``additive terms only'' is a
separate additive-only refit. Contrast rows use donor partition seed $0$; RMS entries are medians over captures. Predicting $0$ has an error
equal to the RMS of the executed $T_C$ over these held-out joint rows, a subset of the states behind the median RMS
in the text.
For held-out prediction, fits use only rows whose donors are all training donors; a joint row is scored only when both its donors are held out.
``Worse than $0$'' counts captures whose nominal $95\%$ donor-bootstrap interval for the excess error over predicting
zero lies above zero (fits held fixed). A \emph{Wald envelope} is, per capture, the median over the $256$ joint states
of the largest $|T_C|$ inside the fit's $95\%$ Wald region for $\gamma$ (point estimate plus radius, all rows of the
family); envelope entries are medians over captures, with the percentage taken per capture. It is a working-model
identification width that reflects the design, the misspecified fit and its residual variance, not an observed error
or a bound on the network. Coverage counts captures in which, at every joint state, the executed $|T_C|$ lies inside
that state's envelope.}
\label{tab:scales}
\begin{tabular}{lcc}
\toprule
 & single-site fit & joint fit \\
\midrule
held-out $R^2$ on joint readouts, full cubic & $\RespSJR$ & $\RespFullThreeJJ$ \\
held-out $R^2$ on joint readouts, additive terms only & $\RespAddThreeSJ$ & $\RespAddThreeJJ$ \\
held-out RMS error predicting $T_C$, logits (predicting $0$: $\MfZero$) & $\MfSThree$ & $\MfJThree$ \\
fits predicting $T_C$ worse than $0$ & $\MfSWorse$ & $\MfJWorse$ \\
Wald envelope, logits & $\AdqEnvS$ & $\AdqEnvJ$ \\
Wald envelope, \% of B-arm swing & $\MisS$ & $\MisJ$ \\
coverage of the executed $|T_C|$ & $\AdqCoverUpperS$ & $\AdqCoverUpperJ$ \\
\bottomrule
\end{tabular}
\end{table}

\paragraph{Design details.} The published pairs combine the S-Inhibition heads 7.3, 7.9, 8.6 and 8.10 with the Name Movers 9.6, 9.9
and 10.0. The control pairs combine the early heads 0.5, 1.2, 2.9 and 3.4, which have no claimed IOI role, with the same Name
Movers.

\paragraph{The executed contrast in detail.} Across the captures on the three base prompts, the interquartile range of
the RMS of $T_C$ is $\AdqRmsTLo$--$\AdqRmsTHi$ logits. Normalising each capture's RMS by its mean absolute per-donor effect at
$\alpha\approx1$ (for each donor, the mean of the $B$-arm readouts at $\alpha=5/6$ and $7/6$ minus the clean readout;
median $\AdqEffectAlphaOne$ logits) gives a median percentage of $\AdqRmsTPctEffect\%$.
The largest single $|T_C|$ in a capture is a median $\AdqMaxTPctEffect\%$ of that effect, and about $\MfMaxAbsMax$ logits at the maximum.

\begin{figure}[h]
\centering
\includegraphics[width=\textwidth]{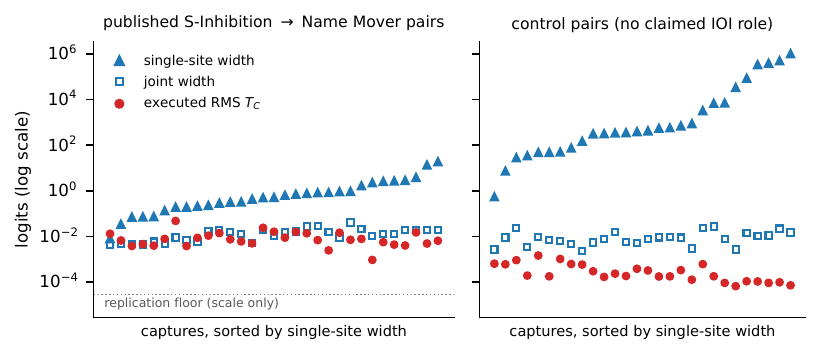}
\caption{The working model cannot stand in for the executed contrast (exploratory; three base prompts). Each horizontal
position is one capture, published S-Inhibition $\to$ Name Mover pairs on the left and controls on the right, sorted by
single-site width. Triangles and open squares are the single-site and joint working-model widths, each a median over
joint states; circles are the RMS of the executed $T_C$ (log scale, logits). For published heads the median
single-site width is $\AdqEnvSOverRmsT$ times the median executed RMS. The widths come from a misspecified model and are
neither observed errors nor calibrated bounds. The dotted line, the median floor of the replication's published
captures, is only a scale reference.}
\label{fig:scales}
\end{figure}

\paragraph{The working model and its chart.} The working model of \S\ref{sec:ioi} uses a fixed, weight-derived chart
motivated by the published claim that S-Inhibition heads act on the Name Movers' query. The upstream coordinate $a$ is
the S-Inhibition head's output along the leading singular direction of $W_O^{\mathrm{SI}}W_Q^{\mathrm{NM}}$, the
composition of its output weights with the Name Mover's query weights. The downstream coordinate $b$ is the Name
Mover's output along its direct effect on the logit difference. Neither direction uses intervention states or
readouts. The working model is the cubic class in this chart, with additive terms in $a$ or $b$ alone and cross
coefficients $\gamma$ on $ab$, $a^2b$ and $ab^2$. Under it $T_C(u,v)=D(u,v)^\top\gamma$, where $D$ is the four-point
contrast of the cross monomials, and the additive terms cancel. On the pinned arm every contrast vanishes, and on the
recompute arm of an affine edge the cross terms reduce to functions of $a$ alone, which the additive terms absorb.
Single-site data therefore inform $\gamma$ only through the scatter of $b$ about the fitted edge. After residualising
the cross terms against the additive terms, we compare the smallest eigenvalues of the per-observation design Grams
in this shared chart. The joint eigenvalue is far larger than the single-site eigenvalue. Their joint/single-site
ratio is the \emph{design gap}; it is a property of the design, not Fisher information about the network.

\paragraph{The working model in detail.} The median edge $R^2$ of the fitted edge is $\QCEdgeRTwo$, so single-site data
inform the cross coefficients only through departures from the affine edge; the low edge $R^2$ does not by itself show that the scatter is small, and the relevant scale comparison is the ratio $\GrpIOIEdge$ of the standard deviation of $b$ on the patch-$A$ arm to that on the patch-$B$ arm. The design gap, whose median in the shared chart is $\QCInfoRatio$, is $\QCInfoRatioOwnChart$ if each family is
standardised separately. The single-site cubic's held-out $R^2$ on joint readouts is negative in at least
$\RespSJNegMin$ of $\MisN$ captures under each of ten donor partitions. For the detection analysis, the ratio of arm-$A$
to arm-$B$ residual variance is a median $\DetArmVarIOI$ for published heads and $\DetArmVarCtrl$ for controls, which
contradicts the equal-variance assumption of the pooled covariance.

\paragraph{Covariance sensitivity and the joint test.} Under a pooled covariance, single-site evidence rejects $\gamma=0$ in
$\QCRejS$ published captures against $\QCCtrlRejS$ controls. Under five alternative covariances controls reject in
$\DetCtrlRobLo$--$\DetCtrlRobHi$ of $\QCCtrlN$ and published heads in $\DetIOIRobLo$--$\DetIOIRobHi$ of $\QCN$. These are
covariance-sensitivity calculations for a misspecified model, not a validated calibration. The replication shows the same
sensitivity to the covariance estimator (Appendix~\ref{app:replication}). The joint fit, with the marginal-sum donor covariance
$V_{\mathrm{up}}+V_{\mathrm{down}}$ (which omits the intersection term and cross-role donor dependence; its calibration is
unknown), rejects in $\DetIOIJTwo$ published and $\DetCtrlJTwo$ control captures.

\paragraph{Estimands of the two tests.} The single-site and joint tests of \S\ref{sec:ioi} need not target the same
coefficients. Correct specification of the cubic model on both supports guarantees a common $\gamma$. With the
misspecification shown in Table~\ref{tab:scales}, each fit instead approximates the network on its own support: near the
two arms for single-site evidence and over the filled region for joint evidence. Sharing a chart therefore does not
guarantee a common estimand. With the pooled working-model covariance $V_S+V_J$, the single-site estimate $\gamma_S$ is
inconsistent with the joint estimate $\gamma_J$ in $\DetInconsistent$ single-site rejections ($\DetInconsistentAll$
overall), and the joint test's nominal plug-in power at $\gamma_S$ rounds to $\DetPowerMin$: had $\gamma_S$ been the joint
fit's target, the joint test would almost always have rejected.

\paragraph{Working-model identification widths.} The widths (computed from nominal Wald regions) of single-site working-model fits vary with degree, covariance and prompt.
Percentages are medians over captures (all $\MisN$ published captures, or those of one prompt where stated)
of each capture's single-site envelope divided by its B-arm swing and expressed as a percentage. By degree: $\EnvSDtwo$ (degree $2$), $\MisS$ ($3$), $\EnvSDfour$ ($4$); the single-site
envelope exceeds the joint one in every capture at every degree, and the smallest ratio of medians over degrees is
$\EnvRatioMin$. By covariance at degree $3$: pooled $\MisS$, HC0 (heteroskedasticity-consistent) $\AdqEnvSHCzero$,
donor-clustered $\AdqEnvSCluster$. By prompt: $\AdqSeedZeroEnvS$, $\AdqSeedOneEnvS$ and $\AdqSeedTwoEnvS$; the
single-site envelope exceeds the whole swing in $\AdqSeedZeroOverOne$, $\AdqSeedOneOverOne$ and $\AdqSeedTwoOverOne$
captures of the three prompts.

Each pointwise envelope is the sum of a centre $|D^\top\hat\gamma|$ and a Wald radius. Taking medians over
joint states and then captures gives centres of $\AdqCentreS$ logits single-site and $\AdqCentreJ$ joint,
and radii of $\AdqRadiusS$ and $\AdqRadiusJ$ logits, respectively. Coverage of the executed $|T_C|$
at all $256$ joint states, under $|T_C|\le|D^\top\hat\gamma|+\text{radius}$: single-site $\AdqCoverUpperSTwo$,
$\AdqCoverUpperS$, $\AdqCoverUpperSFour$ at degrees $2$--$4$; joint $\AdqCoverUpperJTwo$, $\AdqCoverUpperJ$,
$\AdqCoverUpperJFour$. Under $|T_C-D^\top\hat\gamma|\le\text{radius}$: single-site $\AdqCoverIntervalS$ and joint
$\AdqCoverIntervalJ$ at degree $3$. The B-arm swing is a median $\AdqSwingOverEffect$ times the mean absolute per-donor
effect at $\alpha\approx1$.

\paragraph{Controls.} Control head pairs share the Name Movers, prompts and donors with the published pairs. The
exploratory captures did not measure the numerical floor, and the replication did not resolve the control family
under the registered floor rule (Appendix~\ref{app:replication}), so these data support no published/control
transport ratio and no above-control conclusion. Between-prompt and within-prompt standard deviations of $\log$ RMS
$T_C$ were $\AdqSdBetweenPrompt$ and $\AdqSdWithinPrompt$; they informed the replication design.

\paragraph{The edge in the query chart.} In this appendix and in \S\ref{sec:ioi}, the edge strength $s$ of
\S\ref{sec:practical} is computed on all single-site rows of a capture, not only on training rows. The median edge
strength of the published pairs is $\QCEdgeVsCtrl\times$ that of the control pairs (ratio of capture medians). With source heads as units, the four published and four control heads separate completely
(exact one-sided Mann--Whitney $p=\HeadEdgeExactP$, the smallest attainable). Each head's value first averages
$\log_{10}s$ over prompts whose capture passed the screen, then over its Name Movers. This is a descriptive
conditional permutation summary, since the heads share Name Movers, prompts and the model. It is
consistent with query mediation and is not a test of it. The control source heads also sit in earlier layers (0--3) than the published ones (7--8), so this separation does not distinguish claimed role from depth.

\paragraph{Head groups, with heads as units.} We repeat the procedure on previous-token $\to$ induction heads and on
the control group. Induction heads read previous-token information through keys; the final-position query chart and final-position patch do not test that
mechanism, so this group is a comparison of designs, not a second query-mediated circuit. The median edge strength is $\GrpIOIEdge$ for the published IOI pairs, $\GrpIndEdge$ for the
induction group and $\GrpCtrlEdge$ for controls, and the design gap runs the other way: $\GrpIOIInfoRatio$,
$\GrpIndInfoRatio$ and $\GrpCtrlInfoRatio$. The gap is largest for the controls, whose edge is flattest, so it is a
property of the design, not evidence about IOI; the association is largely structural, because a small patch-$A$
spread of $b$ reduces single-site cross-term information. With source heads as units ($n=\HeadRhoN$), the rank
correlation between edge and gap is $\HeadRho$ (exact permutation $p=\HeadRhoP$ one-sided, $\HeadRhoPTwo$ two-sided), a
descriptive conditional permutation summary, since heads share target heads, prompts and the model. Configuration-level
$p$-values would wrongly treat captures that share heads and prompts as independent. For the induction group,
single-site evidence rejects $\gamma=0$ in $\AdqIndRejSHomo$ captures under a pooled covariance and in
$\AdqIndRejSLo$--$\AdqIndRejSHi$ of $\GrpIndN$ under the alternative covariances; the marginal-sum joint test rejects in
$\AdqIndRejJTwo$. 

Table~\ref{tab:ident} returns to the unembedding chart for the eight GPT-2 and Qwen3.5-9B cells,
reporting design information for the cross coefficients.

\begin{table}[!htbp]
\centering\small
\begin{tabular}{lccccc}
\toprule
model $\times$ task & design info., $S$ & design info., $J$ & SE ratio $S/J$ & flip rate $S$ & flip rate $J$ \\
\midrule
GPT-2 $\times$ IOI          & $\IdGPTIOIInfoS$  & $\IdGPTIOIInfoJ$  & $\IdGPTIOISeRatio$  & $\IdGPTIOIFlipS$  & $\IdGPTIOIFlipJ$ \\
GPT-2 $\times$ induction    & $\IdGPTIndInfoS$  & $\IdGPTIndInfoJ$  & $\IdGPTIndSeRatio$  & $\IdGPTIndFlipS$  & $\IdGPTIndFlipJ$ \\
GPT-2 $\times$ capital      & $\IdGPTCapInfoS$  & $\IdGPTCapInfoJ$  & $\IdGPTCapSeRatio$  & $\IdGPTCapFlipS$  & $\IdGPTCapFlipJ$ \\
GPT-2 $\times$ greater-than & $\IdGPTGtInfoS$   & $\IdGPTGtInfoJ$   & $\IdGPTGtSeRatio$   & $\IdGPTGtFlipS$   & $\IdGPTGtFlipJ$ \\
GPT-2 $\times$ agreement    & $\IdGPTAgrInfoS$  & $\IdGPTAgrInfoJ$  & $\IdGPTAgrSeRatio$  & $\IdGPTAgrFlipS$  & $\IdGPTAgrFlipJ$ \\
Qwen3.5-9B $\times$ IOI     & $\IdQwenIOIInfoS$ & $\IdQwenIOIInfoJ$ & $\IdQwenIOISeRatio$ & $\IdQwenIOIFlipS$ & $\IdQwenIOIFlipJ$ \\
Qwen3.5-9B $\times$ induction & $\IdQwenIndInfoS$ & $\IdQwenIndInfoJ$ & $\IdQwenIndSeRatio$ & $\IdQwenIndFlipS$ & $\IdQwenIndFlipJ$ \\
Qwen3.5-9B $\times$ capital & $\IdQwenCapInfoS$ & $\IdQwenCapInfoJ$ & $\IdQwenCapSeRatio$ & $\IdQwenCapFlipS$ & $\IdQwenCapFlipJ$ \\
\bottomrule
\end{tabular}
\caption{Identification of the cross coefficients in the unembedding chart, support-matched captures; entries are
medians over captures. \emph{Design info.} is the smallest eigenvalue of the per-observation residualised design Gram
of the cross block given the additive block: a property of the design, not Fisher information about the network.
The standard-error (SE) ratio is $\|\sqrt{\operatorname{diag}V_S}\|_2/\|\sqrt{\operatorname{diag}V_J}\|_2$,
the ratio of the Euclidean norms of the marginal standard-error vectors in this chart. The flip rate
is the fraction of $40$ random complementary row-half splits on which the fitted $ab$ coefficient changes sign; it is
descriptive and parameterisation-dependent.}
\label{tab:ident}
\end{table}

% ===================================================================== M ====
\clearpage
\section{Related work in full}
\label{app:related}

\paragraph{Algebraic experimental design.} For a finite design, $I(R_Q)$ is the design ideal and the rank of the
evidence map is the Hilbert function of the quotient; aliasing is exactly $N(Q,\mathcal P_{\le d})$. This calculus is
due to \citet{pistone1996groebner}, developed in \citet{pistone2001algebraic} and surveyed by
\citet{maruri2012algebraic}. The algebra is inherited from these works. Our contribution is the modelling step: under patch-and-recompute the network,
not the analyst, produces the design points, which makes $R_Q$ a measurable property of a trained model. Recent work
asks interpretability claims to state their intervention class and identification assumptions
\citep{joshi2026causality,lin2026disclose}.

\paragraph{Comparison of experiments, and one-at-a-time designs.} Ordering experiments by informativeness is due to
\citet{blackwell1953equivalent}. For Gaussian linear experiments with known nonsingular noise covariances and a common
mean parameter, Blackwell informativeness is equivalent to Loewner order of the information matrices
\citep{hansen1974comparison}, and \citet{pukelsheim2006optimal} develops the design theory of Loewner-incomparable
pairs. Our resolving preorder coarsens that order to compare only which directions are resolved within $\mathcal F$. Single-site patching
under clamping is a one-at-a-time design, whose inability to estimate interactions is textbook material
\citep{czitrom1999ofat}; the recompute semantics differ from it only through the realised edge (\S\ref{sec:predict}).

\paragraph{Causal mediation.} Patch-and-recompute measures a total effect; clamping measures a controlled direct effect.
Mediation analysis states the assumptions under which each is identified
\citep{pearl2001direct,robins1992identifiability,vanderweele2015explanation}.
\citet{vanderweele2014fourway} gives the decomposition that separates mediation from interaction. \citet{vig2020gender} brought mediation analysis to
language models. \citet{kekic2025joint} identify joint mean effects from observational and single-variable
interventional distributions under a pairwise-confounded additive structural model and equal action supports across
regimes. Our range checks and executed contrasts diagnose related obstacles but do not establish those hypotheses.
\citet{vaidyanathan2026curse} decompose the patching estimand and expose hidden mediator interactions; that concerns
one chosen intervention's effect, not a family's resolving power.

\paragraph{Mechanistic interpretability.} We ask what a body of intervention evidence can resolve.
\citet{meloux2025everything} establish that mechanistic explanations are not unique, and
\citet{venkatesh2026steering} exhibit large equivalence classes of steering vectors. On intervention semantics,
\citet{makelov2024illusion} show that subspace patching can act through dormant pathways. \citet{geiger2025causal}
supply a general language for describing interventions, and \citet{eslami2026acif} studies separating intervention
families for structural causal models. Positive guarantees address other targets: \citet{asiaee2026cif} certifies a
chosen estimand under a stated intervention distribution, and \citet{dhor2026spectral} prove identifiability for a
model-intrinsic spectral primitive, asking what property of the model is recoverable. 

Complementary evaluations concern intervention success and circuit variability.
\citet{bhalla2025unifying} compare interpretability tools by intervention success; \citet{mueller2025mib} benchmark circuit discovery.
\citet{bayatmakou2026many} and \citet{venkatesh2026architecture} study how recovered circuits vary with inputs,
granularity and architecture. 

Our analysis concerns the intervention families used by automated circuit discovery
\citep{conmy2023acdc} and alignment search \citep{geiger2024das}.
Experimental design for choosing interventions \citep{tigas2022interventions} is the constructive counterpart of our negative statement.
Our protocol follows the patching-practice literature
\citep{zhang2024patching,heimersheim2024patching,goldowskydill2023pathpatching}. Self-repair after ablation \citep{mcgrath2023hydra} and the gap between localisation and editing
\citep{hase2023localization} concern what a successful intervention licenses rather than what a family excludes. The
tasks are those of \citet{wang2022ioi}, \citet{olsson2022induction}, \citet{hanna2023greater}, \citet{meng2022locating}
and \citet{finlayson2021causal}.

% ===================================================================== N ====
\section{Reproducibility and use of AI tools}
\label{app:repro}

A claim ledger supplies the reported numerical results and records each number's source artifacts by hash, unit
and aggregation. Verification re-derives the recorded numerical results from the stored artifacts and compares their formatted
values and artifact hashes. The scope and interpretation of each result are checked separately.

The release at
\url{https://github.com/Ankush-kadu/activation-patching-reachable-state-geometry} contains the analysis code, the
intervention captures including the replication's, the claim ledger, the preregistration with its frozen hashes and
the manuscript sources. Verifying the claims and rebuilding the paper need only a CPU; re-collecting captures needs a GPU.

Large language models assisted with code, protocol drafting, algebraic checking, adversarial review, writing and
literature search. Reported measurements come from executed model interventions and from analyses of stored
artifacts, not from numerical values supplied in model responses. Numerical results and algebraic claims were checked with the archived verification procedures.
The authors reviewed the content and take responsibility for it.

% ===================================================================== O ====
\section{Preregistered replication of the IOI transport analysis}
\label{app:replication}

\paragraph{Registration.} The preregistration, the prompt pool, the collection driver and the analysis script were
frozen by hash, with an empty sealed output directory, before any fresh forward pass. The analysis ran as
registered, with no change after the first data were seen.

\paragraph{Design.} GPT-2 small in float32, with the unchanged collector of \S\ref{sec:ioi}. The $\RepTemplates$ IOI templates of \citet{wang2022ioi} each have a first clause differing from every template
used in this paper's previously inspected captures. We converted them to the name order of our captures and drew
three prompts per template from fixed seeds, giving $\RepPrompts$ prompts. Each prompt has the $12$ published and $12$ control head pairs of
\S\ref{sec:ioi} and the same per-capture design (single-site arms, clamped arm, $256$ joint states), $\RepCaptures$
captures in all. The eight templates are the complete evaluation set and the unit of inference; we report finite-set
criteria over them, not population $p$-values.

\paragraph{Summaries.} For each capture let $r_c$ be the RMS of $T_C$ over the tested joint states and
$q_c=100\,r_c/\text{swing}_c$, where $\text{swing}_c$ is that capture's B-arm readout range.
For prompt $p$, $M_p$ and $M_p^{\mathrm{rel}}$ are the medians of $r_c$ and $q_c$ over its screened published captures;
for template $t$, $V_t$ and $V_t^{\mathrm{rel}}$ are the medians of $M_p$ and $M_p^{\mathrm{rel}}$ over its three prompts.

\begin{center}\small
\begin{tabular}{lcccccccc}
\toprule
template & T1 & T2 & T3 & T4 & T5 & T6 & T7 & T8 \\
\midrule
$V_t$ (logits) & $\RepVTOne$ & $\RepVTTwo$ & $\RepVTThree$ & $\RepVTFour$ & $\RepVTFive$ & $\RepVTSix$ & $\RepVTSeven$ & $\RepVTEight$ \\
$V_t^{\mathrm{rel}}$ (\%) & $\RepVrelTOne$ & $\RepVrelTTwo$ & $\RepVrelTThree$ & $\RepVrelTFour$ & $\RepVrelTFive$ & $\RepVrelTSix$ & $\RepVrelTSeven$ & $\RepVrelTEight$ \\
\bottomrule
\end{tabular}
\end{center}

\paragraph{Validity gates.}
\begin{description}[leftmargin=1.2em,itemsep=1pt,topsep=1pt,parsep=0pt,font=\normalfont]
\item[(V1)] Re-collecting all $24$ captures of an original seed had to reproduce the stored
readouts to within $10^{-4}$ logits and the state coordinates to a relative $10^{-5}$, with equivalent re-execution
under the original batching; all matched exactly.
\item[(V2)] The single-site, clamped and joint rows were re-executed in
reverse order with batch size $7$ instead of $16$, keeping donor activations and the clean pass fixed; a capture's floor
is the RMS of the resulting change in $T_C$, and a family counts as resolved if at least $90\%$ of its captures exceed
ten times their floor.
\item[(V3)] Identity checks: the joint and clamped arms share upstream coordinates, the joint and $B$
arms share downstream coordinates, and pinned coordinates do not move.
\item[(V4)] At least half of the published captures
and at least five templates remain analysable; $\RepScreenedOut$ captures failed the screen, $\RepErrors$ raised errors,
and every template was analysable.
\item[(V5)] Captures with a B-arm swing below $\MfSwingThr$ logits are excluded and
reported; none was.
\end{description}

\paragraph{Registered criteria and outcomes.} H1\textsubscript{abs} passed: $V_t<\MfMargin$ logits was required in
at least $7$ of $8$ templates and observed in $\RepHAbsCount$ (median $\RepVMed$; range $\RepVMin$--$\RepVMax$).
H1\textsubscript{rel} passed: $V_t^{\mathrm{rel}}<1\%$ was required in at least $7$ of $8$ and observed in $\RepHRelCount$ (median $\RepVrelMed\%$; range $\RepVrelMin$--$\RepVrelMax\%$). Both margins were fixed after the exploratory analysis and
are disclosed as such.

H2 assesses the published/control comparison, grouping screened captures by prompt $p$ and Name Mover $n$. In a group with both families present and a positive
control median, $d_{p,n}$ is the log of the median published $r_c$ minus the log of the median control $r_c$; $d_p$ is
the median of the defined $d_{p,n}$ over the prompt's groups and $d_t$ the median of the defined $d_p$ over the
template's prompts. H2 requires $d_t>0$ in at least $7$ of $8$ templates, a positive median of $d_t$, and both
families resolved under V2; a template without a defined $d_t$ does not count. Published contrasts were resolved in $\RepPubResolved$
captures (median RMS $\RepPubRmsMed$ logits; median floor $\RepFloorMed$) but control contrasts in only $\RepCtlResolved$
(median RMS $\RepCtlRmsMed$; median floor $\RepFloorMedCtl$). Under the registered floor rule H2 is floor-limited: we
report no calibrated published/control ratio and no above-control finding. The registered decision rule upgrades the section
only if every criterion passes, so the transport result stays exploratory with its smallness replicated as
registered.

\paragraph{Secondary, descriptive.} Across screened published captures, $q_c$ has maximum $\RepPubRelMax\%$ and 90th
percentile $\RepPubRelPninety\%$; these summarise the RMS over the joint states, not the largest $|T_C|$ at a single
state.

The single-site cubic's held-out prediction error for $T_C$ exceeds that of predicting zero in
$\RepSSixWorse$ published captures, by point error. Single-site and joint Wald envelopes cover the executed $|T_C|$ at all joint states
in $\RepSSevenS$ and $\RepSSevenJ$ published captures. Single-site evidence rejects $\gamma=0$ in $\RepPubHomo$ published
and $\RepCtlHomo$ control captures under a pooled covariance, while controls reject in $\RepCtlRobLo$--$\RepCtlRobHi$ of
$\RepCtlN$ under the alternative covariances; $\gamma_S$ is inconsistent with $\gamma_J$ in $\RepInconsistent$ published
captures. The descriptive sensitivity analyses give the same classifications when excluding the template flagged in the registration
for sharing the phrase ``went to'' with an inspected template (T7), or when dropping the screen. These analyses do not
change the registered decisions.

\end{document}